\documentclass[journal]{IEEEtran}

\usepackage{booktabs}
\usepackage{colortbl}
\usepackage[pdftex]{graphicx}
\graphicspath{{../pdf/}}
\usepackage{amsmath}
\usepackage{algorithm}
\usepackage{algorithmic}
\usepackage{array}
\usepackage{multirow}

\usepackage{graphicx}
\ifCLASSOPTIONcompsoc
\usepackage[caption=false, font=normalsize, labelfont=sf, textfont=sf]{subfig}
\else
\usepackage[caption=false, font=footnotesize]{subfig}
\fi
\usepackage{url}
\usepackage{color,xcolor}

\newcommand{\fref}[1]{Fig. \ref{#1}}
\newcommand{\sref}[1]{Section \ref{#1}}
\newcommand{\tref}[1]{TABLE \ref{#1}}
\newcommand{\eref}[1]{Eq. (\ref{#1})}

\makeatletter

\def\ps@IEEEtitlepagestyle{
	\def\@oddfoot{\mycopyrightnotice}
	\def\@evenfoot{}
}
\def\mycopyrightnotice{
	{\footnotesize 
		
		\begin{minipage}{\textwidth}\ \\[12pt] \centering
			
			\copyright 2020 IEEE. Personal use of this material is permitted. Permission from IEEE must be obtained for all other uses, in any current or future media, including
			reprinting/republishing this material for advertising or promotional purposes, creating new
			collective works, for resale or redistribution to servers or lists, or reuse of any copyrighted
			component of this work in other works.
		\end{minipage}
	
		\hfill} 
	\gdef\mycopyrightnotice{}
}

\begin{document}
\title{Deep Reinforcement Learning for Multi-objective Optimization}

\author{Kaiwen~Li,
        Tao~Zhang,
        and~Rui~Wang
\thanks{This paper is partially supported by the National Natural Science Foundation of China (No. 61773390 and No. 71571187).}  

\thanks{Kaiwen Li, Tao~Zhang and Rui~Wang (corresponding author) are with the College of Systems Engineering, National University of Defense Technology, Changsha 410073, PR China, and also with the Hunan Key Laboratory of Multi-energy System Intelligent Interconnection Technology, HKL-MSI2T, Changsha 410073, PR China. (e-mail: kaiwenli\_nudt@foxmail.com, zhangtao@nudt.edu.cn, ruiwangnudt@gmail.com)}

\thanks{Manuscript received XX XX, 2019; revised XX XX, 2019.}}

\markboth{Journal of IEEE transcations on Cybernetics~2020 DOI: 10.1109/TCYB.2020.2977661}%
{Shell \MakeLowercase{\textit{et al.}}: Deep Reinforcement Learning for Solving the Multi-objective Optimization Problem}

\maketitle

\begin{abstract}
This study proposes an end-to-end framework for solving multi-objective optimization problems (MOPs) using Deep Reinforcement Learning (DRL), that we call DRL-MOA. The idea of decomposition is adopted to decompose the MOP into a set of scalar optimization subproblems. Then each subproblem is modelled as a neural network. Model parameters of all the subproblems are optimized collaboratively according to a neighborhood-based parameter-transfer strategy and the DRL training algorithm. Pareto optimal solutions can be directly obtained through the trained neural network models. In specific, the multi-objective travelling salesman problem (MOTSP) is solved in this work using the DRL-MOA method by modelling the subproblem as a Pointer Network. Extensive experiments have been conducted to study the DRL-MOA and various benchmark methods are compared with it. It is found that, once the trained model is available, it can scale to newly encountered problems with no need of re-training the model. The solutions can be directly obtained by a simple forward calculation of the neural network; thereby, no iteration is required and the MOP can be always solved in a reasonable time. The proposed method provides a new way of solving the MOP by means of DRL. It has shown a set of new characteristics, e.g., strong generalization ability and fast solving speed in comparison with the existing methods for multi-objective optimizations. Experimental results show the effectiveness and competitiveness of the proposed method in terms of model performance and running time. 
\end{abstract}

\IEEEpeerreviewmaketitle

\section{Introduction}
\label{Intro}

\IEEEPARstart{M}{ulti-objective} optimization problems arise regularly in real-world where two or more objectives are required to be optimized simultaneously. Without loss of generality, an MOP can be defined as follows:

\begin{equation}
    \begin{array}{ll}{\min _{\mathbf{x}}} & {\mathbf{f}(\mathbf{x})=\left(f_{1}(\mathbf{x}), f_{2}(\mathbf{x}), \ldots, f_{M}(\mathbf{x})\right)} \\ {\text { s.t. }} & {\mathbf{x} \in X,}
    \end{array}
    \label{eq:mop}
\end{equation}
where $\mathbf{f}(\mathbf{x})$ consists of $M$ different objective functions and $X \subseteq R_D$ is the decision space. Since the $M$ objectives are usually conflicting with each other, a set of trade-off solutions, termed Pareto optimal solutions, are expected to be found for MOPs. 

Among MOPs, various multi-objective combinatorial optimization problems have been investigated in recent years. A canonical example is the multi-objective travelling salesman problem (MOTSP), where given $n$ cities and $M$ cost functions to travel from city $i$ to $j$, one needs to find a cyclic tour of the $n$ cities, minimizing the $M$ cost functions. This is an NP-hard problem even for the single-objective TSP. The best known exact method, i.e., dynamic programming algorithm, requires a complexity of $\Theta\left(2^{n} n^{2}\right)$ for single-objective TSP. It appears to be much harder for its multi-objective version. Hence, in practice, approximate algorithms are commonly used to solve MOTSPs, i.e., finding near optimal solutions. 

During the last two decades, multi-objective evolutionary algorithms (MOEAs) have proven effective in dealing with MOPs since they can obtain a set of solutions in a single run due to their population based characteristic. NSGA-II \cite{deb2002fast} and MOEA/D \cite{zhang2007moea} are two of the most popular MOEAs which have been widely studied and applied in many real-world applications. The two algorithms as well as their variants have also been applied to solve the MOTSP, see e.g., \cite{ke2013moea, beirigo2016application, peng2009comparison}. 

In addition, several handcrafted heuristics especially designed for TSP have been studied, such as the Lin-Kernighan heuristic \cite{lin1973effective} and the 2-opt local search \cite{johnson1990local}. By adopting these carefully designed tricks, a number of specialized methods have been proposed to solve MOTSP, such as the Pareto local search method (PLS) \cite{angel2004dynasearch}, multiple objective genetic local search algorithm (MOGLS) \cite{jaszkiewicz2002performance} and other similar variants \cite{ke2014simple, cai2014external, cai2018grid}. More other methods and details can be found in this review \cite{lust2010multiobjective}. 

Evolutionary algorithms and/or handcrafted heuristics have long been recognized as suitable methods to handle such problems. However, these algorithms, as iteration-based solvers, have suffered obvious limitations that have been widely discussed \cite{zhang2016adecision,ming2019emo,lust2010multiobjective}. 
First, to find near-optimal solutions, especially when the dimension of problems is large, a large number of iterations are required for population updating or iterative searching, thus usually leading to a long running time for optimization. Second, once there is a slight change of the problem, e.g., changing the city locations of the MOTSP, the algorithm may need to be re-performed to compute the solutions. When it comes to newly encountered problems, or even new instances of a similar problem, the algorithm needs to be revised to obtain a good result, which is known as the \emph{No Free Lunch theorem} \cite{wolpert1997no}. Furthermore, such problem specific methods are usually optimized for one task only.

Carefully handcrafted evolution strategies and heuristics can certainly improve the performance. However, the recent advances in machine learning algorithms have shown their ability of replacing humans as the engineers of algorithms to solve different problems. Several years ago, most people used man-engineered features in the field of computer vision but now the Deep Neural Networks (DNNs) have become the main techniques. While DNNs focus on making \emph{predictions}, Deep Reinforcement Learning (DRL) is mainly used to learn how to make \emph{decisions}. Thereby, we believe that DRL is a possible way of learning how to solve various optimization problems automatically, thus demanding no man-engineered evolution strategies and heuristics. 

In this work, we explore the possibility of using DRL to solve MOPs, MOTSP in specific, in an end-to-end manner, i.e., given $n$ cities as input, the optimal solutions can be \emph{directly} obtained through a forward propagation of the trained network. The network model is trained through the trial and error process of DRL and can be viewed as a black-box heuristic or a meta-algorithm \cite{nazari2018reinforcement} with strong learned heuristics. Because of the exploring characteristic of DRL training, the obtained model can have a strong generalization ability, that is, it can solve the problems that it never saw before. 

This work is originally motivated by several recent proposed Neural Network-based single-objective TSP solvers. \cite{vinyals2015pointer} first proposes a Pointer Network that uses attention mechanism \cite{bahdanau2014neural} to predict the city permutation. This model is trained in a supervised way that requires enormous TSP examples and their optimal tours as training set. It is hard for use and the supervised training process prevents the model from obtaining better tours than the ones provided in the training set. 
To resolve this issue, \cite{bello2016neural} adopts an Actor-Critic DRL training algorithm to train the Point Network with no need of providing the optimal tours.  \cite{nazari2018reinforcement} simplifies the Point Network model and adds dynamic elements input to extend the model to solve the Vehicle Routing Problem (VRP). 
Moreover, the advanced Transformer model is employed to solve the routing problems and proves to be effective as well \cite{kool2018attention, deudon2018learning}. 

The recent progress in solving the TSP by means of DRL is really appealing and inspiring due to its non-iterative yet efficient characteristic and strong generalization ability. However, there are no such studies concerning solving MOPs (or the MOTSP in specific) by DRL-based methods. 

Therefore, this paper proposes to use the DRL method to deal with MOPs based on a simple but effective framework. And the MOTSP is taken as a specific test problem to demonstrate its effectiveness.

The main contributions of our work are as follows:
\begin{itemize}
    \item This work provides a new way of solving the MOP by means of DRL. Some encouragingly new characteristics of the proposed method have been found in comparison with classical methods, e.g., strong generalization ability and fast solving speed.
    \item With a slight change of the problem instance, the classical methods usually need to be re-conducted from scratch, which is impractical for application especially for large-scale problems. In contrast, our method is robust to problem changes. Once the model is trained, it can scale to problems that the algorithm never saw before in terms of the number and locations of the cities of the MOTSP. 
    \item The proposed method requires much lower running time than the classical methods, since the Pareto optimal solutions can be directly obtained by a simple forward propagation of the trained networks without any population updating or iterative searching procedures. 
    \item Empirical studies show that the proposed method significantly outperforms the classical methods especially for large-scale MOTSPs in terms of both convergence and diversity, while requiring much less running time. 
\end{itemize}

It is noted that several papers \cite{hsu2018monas, mossalam2016multi} have introduced the concept of \emph{Multi-objective deep reinforcement learning} in the field of RL. However, they mainly focus on how to apply RL to control a robot with multiple goals, such as controlling a mountain car or controlling a submarine searching for treasures, as investigated in \cite{hsu2018monas}. These studies are not explicitly proposed to deal with mathematical optimization problems like \eref{eq:mop}, and thus is out of the scope of this study.

The structure of the paper is organized as follows: \sref{framework} introduces the general framework of DRL-MOA that describes the idea of using DRL for solving MOPs. And \sref{model} elaborates the detailed modelling and training process of solving the specific MOTSP problem by means of the proposed DRL-MOA framework. Finally the effectiveness of the method is demonstrated through experiments in \sref{exp} and \sref{result}.

\section{The Deep Reinforcement learning-based multi-objective optimization algorithm (DRL-MOA)}
\label{2}
In this section, we propose to solve the MOP by means of DRL based on a simple but effective framework (DRL-MOA). First, the decomposition strategy \cite{zhang2007moea} is adopted to decompose the MOP into a number of subproblems. Each subproblem is modelled as a neural network. Then model parameters of all the subproblems are optimized collaboratively according to the neighborhood-based parameter transfer strategy and the Actor-Critic \cite{mnih2016asynchronous} training algorithm. In particular, MOTSP is taken as a specific problem to elaborate how to model and solve the MOP based on the DRL-MOA. 
\subsection{General framework}
\label{framework}
\textbf{Decomposition strategy.} \quad 
Decomposition, as a simple yet efficient way to design the multi-objective optimization algorithms, has fostered a number of researches in the community, e.g., Cellular-based MOGA\cite{murata2001CMOGA}, MOEA/D, MOEA/DD \cite{li2015evolutionary}, DMOEA-$\varepsilon \text{C}$ \cite{Jie2017DMOEA} and NSGA-III \cite{deb2014evolutionary}. The idea of decomposition is also adopted as the basic framework of the proposed DRL-MOA in this work. Specifically, the MOP, e.g., the MOTSP, is explicitly decomposed into a set of scalar optimization subproblems and solved in a collaborative manner. Solving each scalar optimization problem usually leads to a Pareto optimal solution. The desired Pareto Front (PF) can be obtained when all the scalar optimization problems are solved. 

\begin{figure}[tbph]
\centering
\includegraphics[width=2.4in]{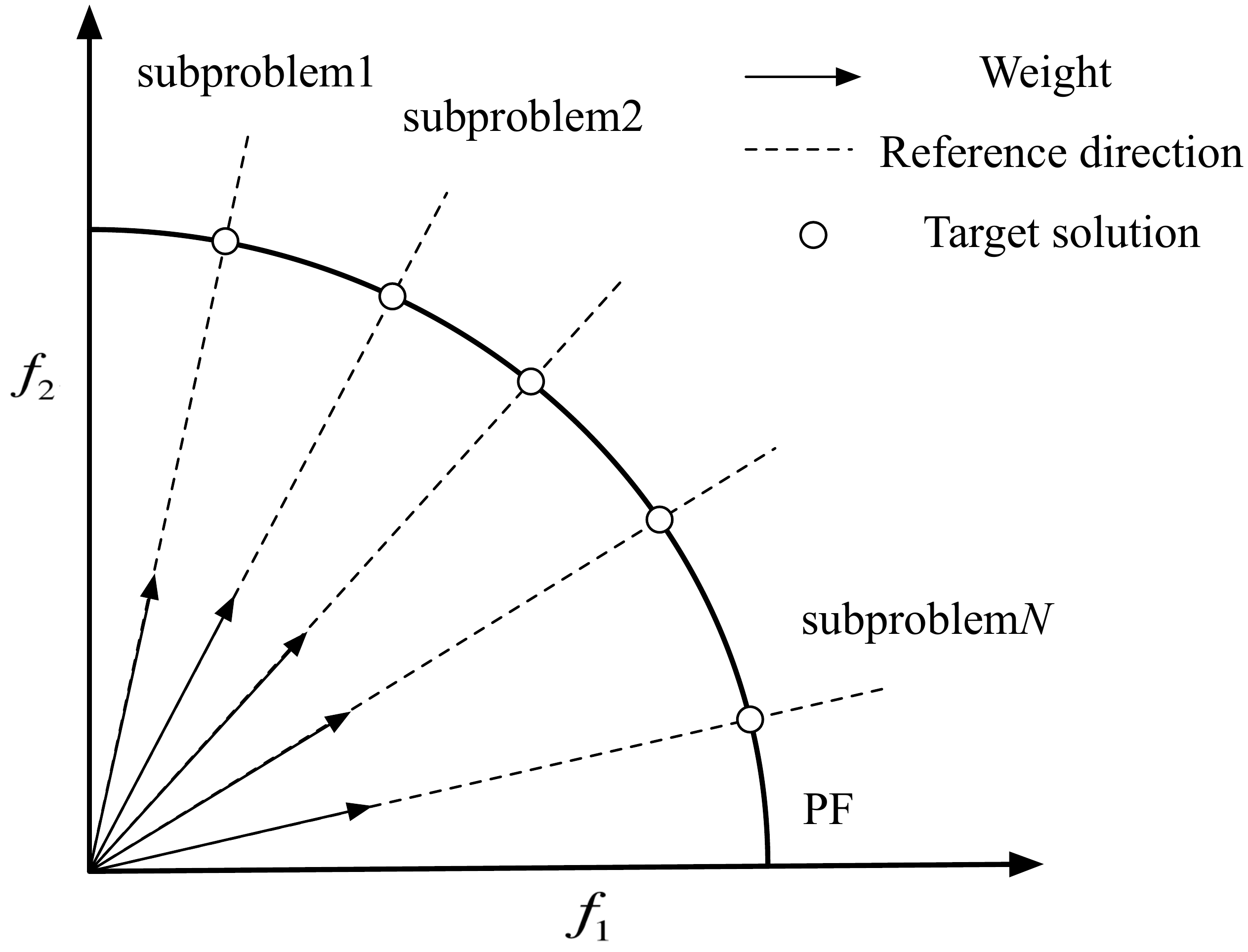}
\caption{Illustration of the decomposition strategy.}
\label{fig:ws}
\end{figure}

In specific, the well-known Weighted Sum \cite{miettinen2012nonlinear} approach is employed. Certainly, other scalarizing methods can also be applied, e.g., the Chebyshev and the penalty-based boundary intersection (PBI) method \cite{wang2018localized,wang2016decomposition}. First a set of uniformly spread weight vectors $\lambda^{1}, \ldots, \lambda^{N}$ is given, e.g., $(1,0), (0.9,0.1), \ldots, (0,1)$ for a bi-objective problem, as shown in \fref{fig:ws}. Here $\lambda^{j}=(\lambda_{1}^{j}, \ldots, \lambda_{M}^{j})^{T}$, where $M$ represents the number of objectives. Thus the original MOP is converted into $N$ scalar optimization subproblems by the Weighted Sum approach. The objective function of the $j_{th}$ subproblem is shown as follows \cite{zhang2007moea}:
\begin{equation}
    \operatorname{minimize} g^{ws}(x | \lambda_i^j)=\sum_{i=1}^{M} \lambda_{i}^j f_{i}(x)
    \label{eq:ws}
\end{equation}

Therefore, the PF can be formed by the solutions obtained by solving all the $N$ subproblems.

\textbf{Neighborhood-based parameter-transfer strategy.} \quad  To solve each subproblem by means of DRL, the subproblem is modelled as a neural network. Then the $N$ scalar optimization subproblems are solved in a collaborative manner by the neighborhood-based parameter-transfer strategy, which is introduced as follows. 

According to \eref{eq:ws} it is observed that two neighbouring subproblems could have very close optimal solutions \cite{zhang2007moea} as their weight vectors are adjacent. Thus, a subproblem can be solved assisted by the knowledge of its neighboring subproblems.  
\begin{figure}[tbph]
\centering
\includegraphics[width=2.2in]{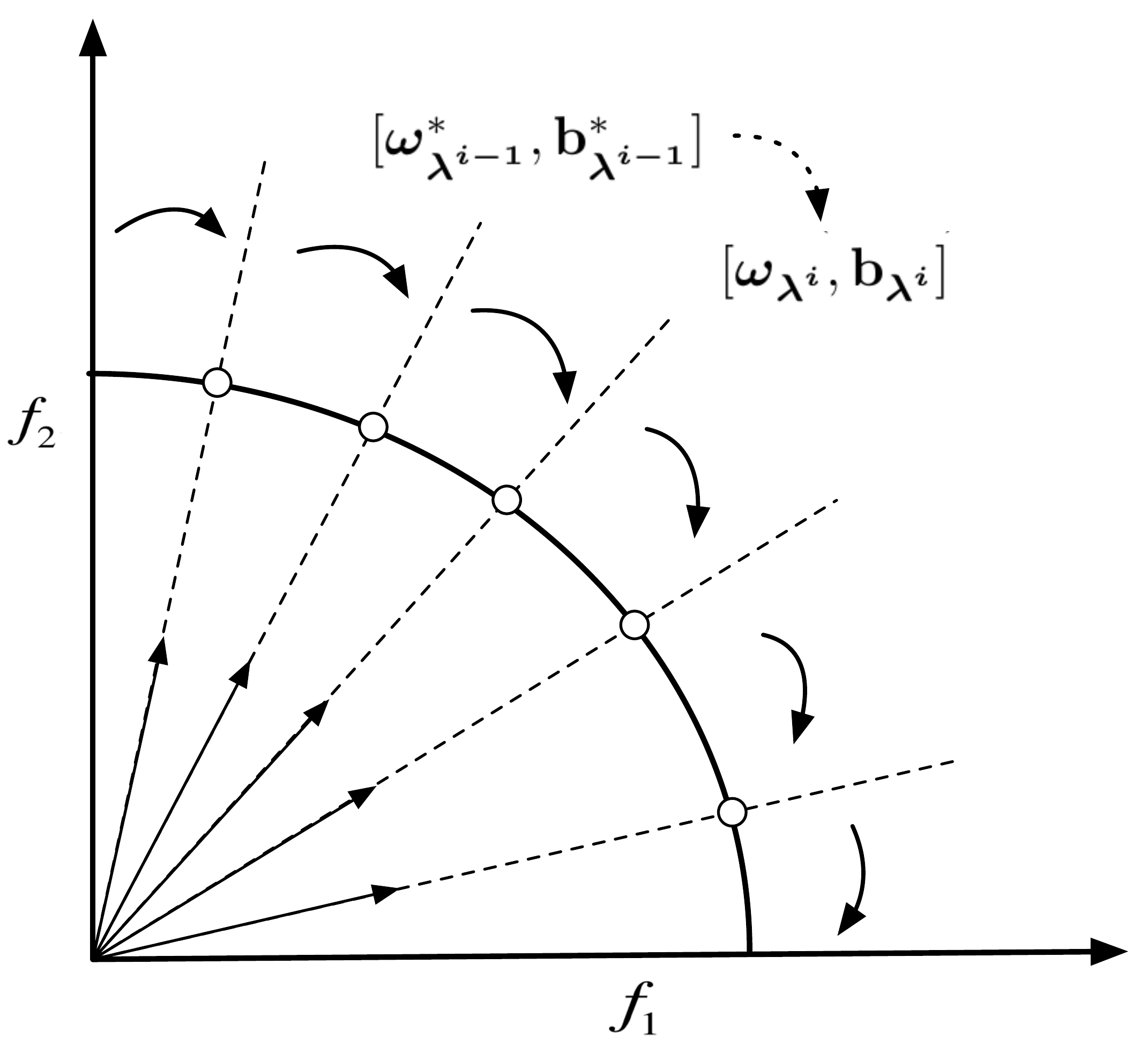}
\caption{Illustration of the parameter-transfer strategy.}
\label{fig:parameter}
\end{figure}

Specifically, as the subproblem in this work is modelled as a neural network, the parameters of the network model of the ${(i-1)}_{th}$ subproblem can be expressed as $[\boldsymbol{\omega}_{\boldsymbol{\lambda^{i-1}}},\mathbf{b}_{\boldsymbol{\lambda^{i-1}}}]$. Here, $[\boldsymbol{\omega}^*,\mathbf{b}^*]$ represents the parameters of the neural network model that have been optimized already and $[\boldsymbol{\omega},\mathbf{b}]$ represents the parameters that are not optimized yet. Assume that the ${(i-1)}_{th}$ subproblem has been solved, i.e., its network parameters have been optimized to its near optimum. Then the best network parameters $[\boldsymbol{\omega}_{\boldsymbol{\lambda^{i-1}}}^*,\mathbf{b}_{\boldsymbol{\lambda^{i-1}}}^*]$ obtained in the ${(i-1)}_{th}$ subproblem are set as the starting point for the network training in the $i_{th}$ subproblem. Briefly, the network parameters are transferred from the previous subproblem to the next subproblem in a sequence, as depicted in \fref{fig:parameter}. The neighborhood-based parameter-transfer strategy makes it possible for the training of the DRL-MOA model; otherwise a tremendous amount of time is required for training the $N$ subproblems. 

Each subproblem is modelled and solved by the DRL algorithm and all subproblems can be solved in sequence by transferring the network weights. Thus, the PF can be finally approximated according to the obtained model. Employing the decomposition in conjunction with the neighborhood-based parameter transfer strategy, the general framework of DRL-MOA is presented in Algorithm \ref{alg:general}.

\begin{algorithm}[htb]
\caption{General Framework of DRL-MOA}
\label{alg:general}
\begin{algorithmic}[1]
\REQUIRE {The model of the subproblem $\mathcal{M} = [\boldsymbol{w},\mathbf{b}]$, weight vectors $\boldsymbol{\lambda^{1}, \ldots, \lambda^{N}}$}
\ENSURE {The optimal model $\mathcal{M}^* = [\boldsymbol{w}^*,\mathbf{b}^*]$}
\STATE $[\boldsymbol{\omega}_{\boldsymbol{\lambda^{1}}},\mathbf{b}_{\boldsymbol{\lambda^{1}}}] \leftarrow Random\_Initialize$

\FOR{$i \leftarrow 1:N$}
\IF {$i ==1$}
\STATE $[\boldsymbol{\omega}_{\boldsymbol{\lambda^{1}}}^*,\mathbf{b}_{\boldsymbol{\lambda^{1}}}^*] \leftarrow Actor\_Critic([\boldsymbol{\omega}_{\boldsymbol{\lambda^{1}}},\mathbf{b}_{\boldsymbol{\lambda^{1}}}],g^{ws}(\boldsymbol{\lambda^{1}}))$
\ELSE 
\STATE $[\boldsymbol{\omega}_{\boldsymbol{\lambda^{i}}},\mathbf{b}_{\boldsymbol{\lambda^{i}}}] \leftarrow [\boldsymbol{\omega}_{\boldsymbol{\lambda^{i-1}}}^*,\mathbf{b}_{\boldsymbol{\lambda^{i-1}}}^*]$
\STATE $[\boldsymbol{\omega}_{\boldsymbol{\lambda^{i}}}^*,\mathbf{b}_{\boldsymbol{\lambda^{i}}}^*] \leftarrow Actor\_Critic([\boldsymbol{\omega}_{\boldsymbol{\lambda^{i}}},\mathbf{b}_{\boldsymbol{\lambda^{i}}}],g^{ws}(\boldsymbol{\lambda^{i}}))$
\ENDIF
\ENDFOR
\RETURN $[\boldsymbol{w}^*,\mathbf{b}^*]$
\STATE Given inputs of the $MOP$, the $PF$ can be directly calculated by $[\boldsymbol{w}^*,\mathbf{b}^*]$.
\end{algorithmic}
\end{algorithm}

One obvious advantage of the DRL-MOA is its modularity and simplicity for use. For example, the MOTSP can be solved by integrating any of the recently proposed novel DRL-based TSP solvers \cite{nazari2018reinforcement, kool2018attention} into the DRL-MOA framework. Also, other problems such as VRP and Knapsack problem can be easily handled with the DRL-MOA framework by simply replacing the model of the subproblem. Moreover, once the trained model is available, the PF can be directly obtained by a simple forward propagation of the model. 

The proposed DRL-MOA acts as an outer-loop. The next issue is how to model and solve the decomposed scalar subproblems. Therefore we take the MOTSP as a specific example and introduce how to model and solve the subproblem of MOTSP in the next section.

\subsection{Modelling the subproblem of MOTSP}
\label{model}
To solve the MOTSP, we first decompose the MOTSP into a set of subproblems and solve each one collaboratively based on the foregoing DRL-MOA framework. Each subproblem is modelled and solved by means of DRL. This section introduces how to model the subproblem of MOTSP in a neural network manner. Here, a modified Pointer Network similar to \cite{nazari2018reinforcement} is used to model the subproblem and the Actor-Critic algorithm is used for training. 

\subsubsection{Formulation of MOTSP}
We recall the formulation of an MOTSP. One needs to find a tour of $n$ cities, i.e., a cyclic permutation $\rho$, to minimize $M$ different cost functions simultaneously:
\begin{equation}
     \min z_{k}(\rho)=\sum_{i=1}^{n-1}{c^{k}_{\rho(i), \rho(i+1)}+c^{k}_{\rho(n), \rho(1)}},k=1,\ldots,M
\end{equation}
where $c^{k}_{\rho(i), \rho(i+1)}$ is the $k_{th}$ cost of travelling from city $\rho(i)$ to $\rho(i+1)$. The cost functions may for example correspond to tour length, safety index or tourist attractiveness in practical applications. 
\subsubsection{The model}
In this part, the above problem is modelled using a modified Pointer Network \cite{nazari2018reinforcement}.

\begin{figure}[tbph]
\centering
\includegraphics[width=2.2in]{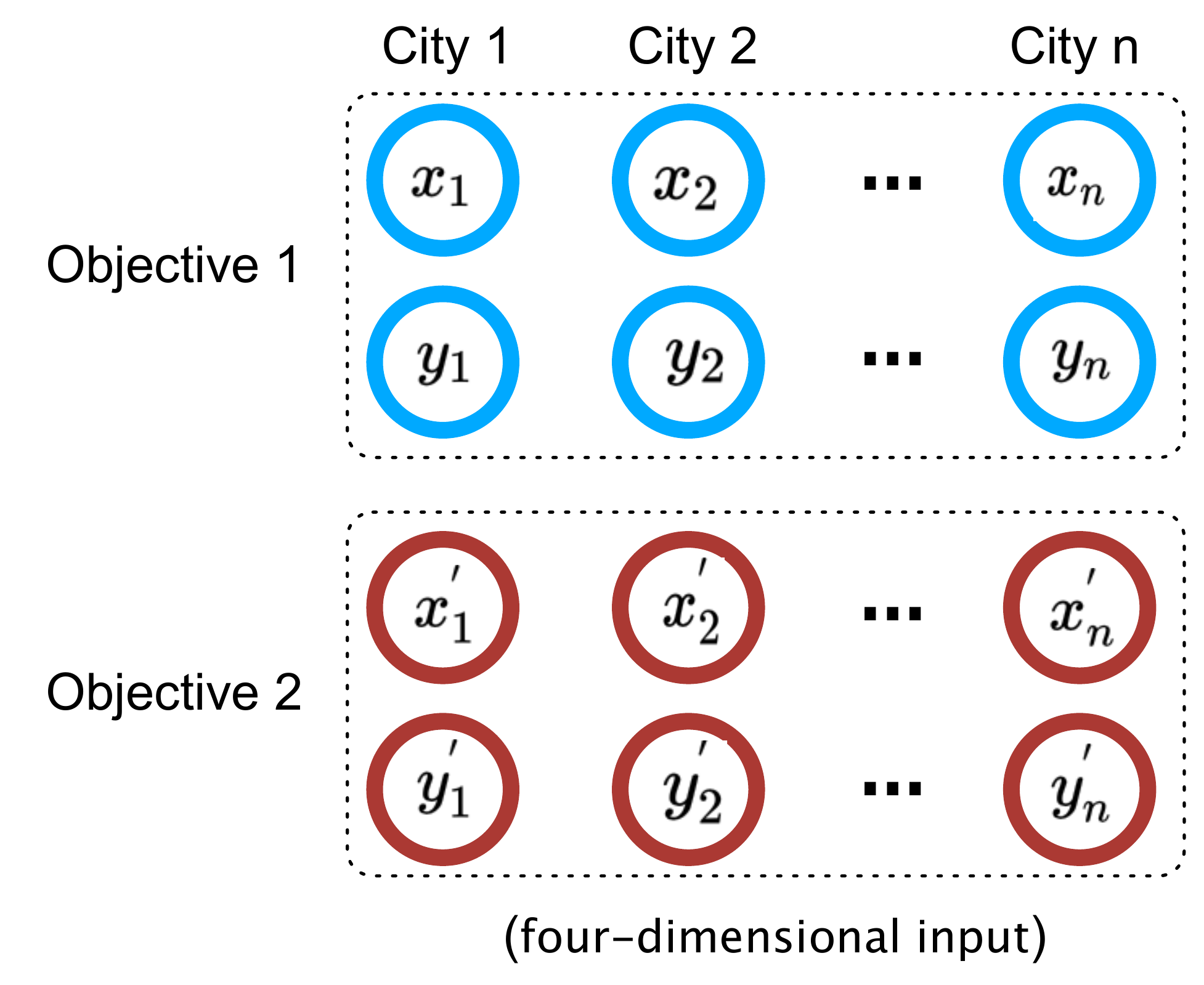}
\caption{Input structures of the neural network model for solving Euclidean bi-objective TSPs.}
\label{fig:input}
\end{figure}

First the input and output structure of the network model is introduced: let the given set of inputs be $X \doteq\left\{s^{i}, i=1, \cdots, n\right\}$ where $n$ is the number of cities. Each $s^{i}$ is represented by a tuple $\left\{s^{i}=\left(s^{i}_1, \cdots, s_{M}^{i}\right) \right\}$. $s^{i}_j$ is the attribute of the $i_{th}$ city that is used to calculate the $j_{th}$ cost function. For instance, $s^{i}_1=(x_i,y_i)$ represents the x-coordinate and y-coordinate of the $i_{th}$ city and is used to calculate the distance between two cities. Taking a bi-objective TSP as an example where both the two cost functions are defined by the Euclidean distance \cite{lust2010multiobjective}, the input structure is shown in \fref{fig:input}. The input is four-dimensional and consists of total $4\times n$ values. Moreover, the output of the model is a permutation of the cities $Y=\left\{ \rho_1, \cdots, \rho_n \right\}$. 

To map input $X$ to output $Y$, the probability chain rule is used: 
\begin{equation}
   P\left(Y|X\right)=\prod_{t=1}^{n} P\left(\rho_{t+1} | \rho_1,\cdots,\rho_t, X_{t}\right).
\label{eq:chain}
\end{equation}

First an arbitrary city is selected as $\rho_1$. At each decoding step $t=1,2,\cdots$, we choose $\rho_{t+1}$ from the available cities $X_t$. The available cities $X_t$ are updated every time a city is visited. In a nutshell, \eref{eq:chain} provides the probability of selecting the next city according to $\rho_1,\cdots,\rho_t$, i.e., the already visited cities.

Then a modified Pointer network similar to \cite{nazari2018reinforcement} is used to model \eref{eq:chain}. Its basic structure is the Sequence-to-Sequence model \cite{sutskever2014sequence}, a recently proposed powerful model in the field of machine translation, which maps one sequence to another. The general Sequence-to-Sequence model consists of two RNN networks, termed encoder and decoder. An encoder RNN encodes the input sequence into a code vector that contains knowledge of the input. Based on the code vector, a decoder RNN is used to decode the knowledge vector to a desired sequence. Thus, the nature of Sequence-to-Sequence model that maps one input sequence to an output sequence is suitable for solving the TSP.

\begin{figure}[tbph]
\centering
\includegraphics[width=3.5in]{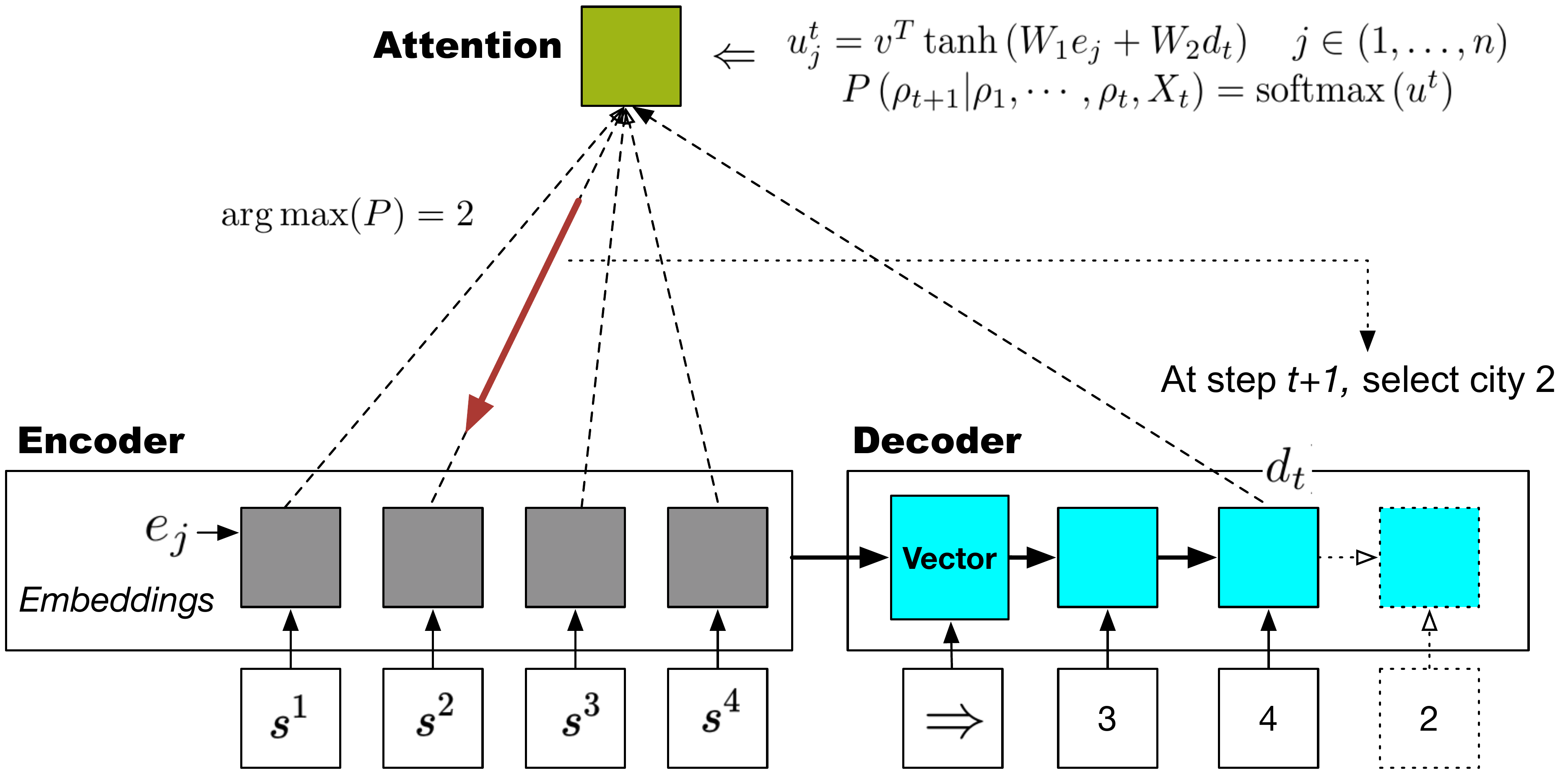}
\caption{Illustration of the model structure. Attention mechanism, in conjunction with Encoder and Decoder, produces the probability of selecting the next city.}
\label{fig:attention}
\end{figure}

In this work, the architecture of the model is shown in  \fref{fig:attention} where the left part is the encoder and the right part is the decoder. The model is elaborated as follows.

\textbf{Encoder.} \quad 
Encoder is used to condense the input sequence into a vector. Since the coordinates of the cities convey no sequential information \cite{nazari2018reinforcement} and the order of city locations in the inputs is not meaningful, RNN is not used in the encoder in this work. Instead, a simple embedding layer is used to encode the inputs to a vector which can decrease the complexity of the model and reduce the computational cost. Specifically, the 1-dimensional (1-D) convolution layer is used to encode the inputs to a \emph{high-dimensional vector} \cite{nazari2018reinforcement} ($d_h$=128 in this paper), as shown in \fref{fig:attention}. The number of in-channels equals to the dimension of the inputs. For example, the Euclidean bi-objective TSP has a four-dimensional input as shown in \fref{fig:input} and thus the number of in-channels is four. And the Encoder finally results to a $n \times d_h$ vector where $n$ indicates the city number. It is noteworthy that the parameters of the 1-D convolution layer are shared amongst all the cities. It means that, no matter how many cities there are, each city shares the same set of parameters which encode the city information to a high-dimensional vector. Thus, the encoder is robust to the number of the cities.

\textbf{Decoder.} \quad 
The decoder is used to unfold the obtained high-dimensional vector, which stores the knowledge of inputs, into the output sequence. Different from the encoder, a RNN is required in the decoder as we need to summarize the information of previously selected cities $\rho_1, \cdots, \rho_t$ so as to make the decision of $\rho_{t+1}$. RNN has the ability of memorizing the previous outputs. In this work we adopt the RNN model of GRU (Gated recurrent unit) \cite{cho2014learning} that has similar performance but fewer parameters than the LSTM (Long Short-Term Memory) which is employed in the original Pointer Network in \cite{nazari2018reinforcement}. It is noted that RNN is not directly used to output the sequence. What we need is the RNN decoder hidden state $d_t$ at decoding step $t$ that stores the knowledge of previous steps $\rho_1, \cdots, \rho_t$. Then $d_t$ and the encoding of the inputs $e_1,\cdots,e_n$ are used together to calculate the conditional probability $P\left(y_{t+1} | \rho_1, \cdots, \rho_t, X_{t}\right)$ over the next step of city selection. This calculation is realized by the attention mechanism. As shown in \fref{fig:attention}, to select the next city at step $t+1$, first we obtain the hidden state $d_t$ through Decoder. In conjunction with $e_1,\cdots,e_n$, the index of next city can be calculated using the attention mechanism.

\textbf{Attention mechanism.} \quad 
Intuitively, the attention mechanism calculates how much every input is relevant in the next decoding step $t$. The most relevant one is given more \emph{attention} and can be selected as the next visiting city. The calculation is as follows:
\begin{equation}
    \begin{array}
{c}{u_{j}^{t}=v^{T} \tanh \left(W_{1} e_{j}+W_{2} d_{t}\right) \quad j \in(1, \ldots, n)} \\ 
{P\left(\rho_{t+1} | \rho_1, \cdots, \rho_t, X_{t}\right)=\operatorname{softmax}\left(u^{t}\right)}
\end{array}
\end{equation}
where $v, W_1, W_2$ are \textit{learnable} parameters. $d_t$ is a key variable for calculating $P\left(\rho_{t+1} | \rho_1, \cdots, \rho_t, X_{t}\right)$ as it stores the information of previous steps $\rho_1, \cdots, \rho_t$. Then, for each city $j$, its $u_{j}^{t}$ is computed by $d_t$ and its encoder hidden state $e_j$, as shown in  \fref{fig:attention}. The softmax operator is used to normalize $u_1^t,\cdots,u_n^t$ and finally the probability for selecting each city $j$ at step $t$ can be finally obtained. The greedy decoder can be used to select the next city. For example, in  \fref{fig:attention}, city 2 has the largest $P\left(\rho_{t+1} | \rho_1, \cdots, \rho_t, X_{t}\right)$ and so is selected as the next visiting city. Instead of selecting the city with the largest probability greedily, during training, the model selects the next city by sampling from the probability distribution.

\subsubsection{Training method}
The model of the subproblem is trained using the well-known Actor-critic method similar to \cite{bello2016neural, nazari2018reinforcement}. However, as \cite{bello2016neural, nazari2018reinforcement} trains the model of single-objective TSP, the training procedure is different for the MOTSP case, as presented in Algorithm \ref{alg:ac}. Next we briefly introduce the training procedure.

Two networks are required for training: ($i$) an actor network, which is exactly the Pointer Network in this work, gives the probability distribution for choosing the next action, and ($ii$) a critic network that evaluates the expected reward given a specific problem sate. The critic network employs the same architecture as the Pointer network's encoder that maps the encoder hidden state into the critic output.  

The training is conducted in an unsupervised way. During the training, we generate the MOTSP instances from distributions $\left\{\Phi_{\mathcal{M}_1},\cdots,\Phi_{\mathcal{M}_M}\right\}$. Here, $\mathcal{M}$ represents different input features of the cities, e.g., the city locations or the security indices of the cities. For example, for Euclidean instances of a bi-objective TSP, $\mathcal{M}_1$ and $\mathcal{M}_2$ are both city coordinates and $\Phi_{\mathcal{M}_1}$ or $\Phi_{\mathcal{M}_2}$ can be a uniform distribution of $[0,1] \times [0,1]$. 

To train the actor and critic networks with parameters $\theta$ and $\phi$, $N$ instances are sampled from $\left\{\Phi_{\mathcal{M}_1},\cdots,\Phi_{\mathcal{M}_M}\right\}$ for training. For each instance, we use the actor network with current parameters $\theta$ to produce the cyclic tour of the cities and the corresponding reward can be computed. Then the policy gradient is computed in line 11 (refer to \cite{konda2000actor} for details of the formula derivation of policy gradient) to update the actor network. Here, $V\left(X_{0}^{n} ; \phi\right)$ is the reward approximation of instance $n$ calculated by the critic network. The critic network is then updated in line 12 by reducing the difference between the true observed rewards and the approximated rewards.

\begin{algorithm}[htb]
\caption{Actor-Critic training algorithm}
\label{alg:ac}
\begin{algorithmic}[2]
\REQUIRE {$\theta, \phi \leftarrow $ initialized parameters given in Algorithm 1}
\ENSURE {The optimal parameters $\theta, \phi$}

\FOR{$iteration \leftarrow 1,2,\cdots$}

\STATE generate $T$ problem instances from $\left\{\Phi_{\mathcal{M}_1},\cdots,\Phi_{\mathcal{M}_M}\right\}$ for the MOTSP.
\FOR{$k \leftarrow 1,\cdots,T$}
\STATE $t \leftarrow 0$
\WHILE{not terminated}
\STATE select the next city $\rho_{t+1}^k$ according to $P\left(\rho_{t+1}^{k} | \rho_{1}^{k},\cdots,\rho_{t}^{k}, X_{t}^{k}\right)$
\STATE Update $X_t^k$ to $X_{t+1}^k$ by leaving out the visited cities.
\ENDWHILE
\STATE compute the reward $R^k$
\ENDFOR
\STATE $d \theta \leftarrow \frac{1}{N} \sum_{k=1}^{N}\left(R^{k}-V\left(X_{0}^{k} ; \phi\right)\right) \nabla_{\theta} \log P\left(Y^{k} | X_{0}^{k}\right)$
\STATE $d \phi \leftarrow \frac{1}{N} \sum_{k=1}^{N} \nabla_{\phi}\left(R^{k}-V\left(X_{0}^{k} ; \phi\right)\right)^{2}$
\STATE $\theta \leftarrow \theta + \eta d\theta$ 
\STATE $\phi \leftarrow \phi + \eta d\phi $
\ENDFOR

\end{algorithmic}
\end{algorithm}

Once all of the models of the subproblems are trained, the Pareto optimal solutions can be directly output by a simple forward propagation of the models. Time complexity of a forward calculation of Encoder is $O(d_hn)$. And time complexity of a forward calculation of Decoder is $O(d_h^2n)$, where $O(d_h^2)$ is the approximated time complexity of the RNN. Thus the approximated time complexity of using DRL-MOA for solving the MOTSP is $O(Nnd_h^2)$ where N is the number of subproblems. As a forward propagation of the Encoder-Decoder neural network can be quite fast, the solutions can be always obtained within a reasonable time. 

\section{Experimental Setups}
\label{exp}
The proposed DRL-MOA is tested on bi-objective TSPs. All experiments are conducted on a single GTX 2080Ti GPU. The code is written in Python and is publicly available \footnote{https://github.com/kevin031060/RL\_TSP\_4static} to reproduce the experimental results and to facilitate future studies. Meanwhile, all experiments of the compared MOEAs are conducted on the standard software platform PlatEMO\footnote{ http://bimk.ahu.edu.cn/index.php?s=/Index/Software/index.html} \cite{Tian2017PlatEMO} which is written in MATLAB. The compared MOGLS is written in Python\footnote{https://github.com/kevin031060/Genetic\_Local\_Search\_TSP}. All of the compared algorithms are run on the Intel 16-Core i7-9800X CPU with 64GB memory. 
\subsection{Test Instances}
The considered bi-objective TSP instances are described as follows \cite{lust2010multiobjective}:

\textbf{Euclidean instances}: Euclidean instances are the commonly used test instances for solving MOTSP \cite{lust2010multiobjective}. Intuitively, both the cost functions are defined by the Euclidean distance. The first cost is defined by the distance between the real coordinates of two cities $i,j$. The second cost of travelling from city $i$ to city $j$ is defined by another set of \emph{virtual} coordinates that used to calculate another objective. Thus the input is four-dimensional as shown in \fref{fig:input}. 

  \textbf{Mixed instances}: In order to test the ability of our model that can adapt to different input structures, Mixed instances with three-dimensional input are tested. Here, the first cost function is still defined by the Euclidean distance between two points that represents the real city location, which is a two-dimensional input with x-coordinate and y-coordinate. Moreover, the second cost of travelling from city $i$ to $j$ is defined by a one-dimensional input. This one-dimensional input of city $i$ can be interpreted as the altitude of city $i$. Thus the objective is to minimize the altitude variance when travelling between two cities. A smoother tour can be obtained with less altitude variance, therefore, we can reduce the fuel cost or improve the comfort of the journey. Thereby, Mixed instances have a three-dimensional input.

\textbf{Training set}:
As an unsupervised learning method, only the model input and the reward function are required during the training process, with no need of the best tours as the labels. Euclidean instances with four-dimensional input and Mixed instances with three-dimensional input are generated as the training set. They are all generated from a uniform distribution of $[0,1]$.

\textbf{Test set}:
The standard TSP test problems kroA and kroB in the TSPLIB library \cite{reinelt1991tsplib} are used to construct the Euclidean test instances kroAB100, kroAB150 and kroAB200 which are commonly used MOTSP test instances \cite{lust2010multiobjective, cai2018grid}. kroA and kroB are two sets of different city locations and used to calculate the two Euclidean costs. For Mixed test instances, randomly generated 40-, 70-, 100-, 150- and 200-city instances are constructed.

In this work, the model is trained on 40-city MOTSP instances and it is used to approximate the PFs of 40-, 70-, 100-, 150- and 200-city test instances.

\subsection{Parameter settings of model and training }
Most parameters of the model and training are similar to that in \cite{nazari2018reinforcement} which can solve single-objective TSPs  effectively. Specifically, the parameter settings of the network model are shown in \tref{parameter:model}. $D_{input}$ represents the dimension of input, i.e., $D_{input} = 4$ for Euclidean bi-objective TSPs. We employ an one-layer GRU RNN with the hidden size of 128 in the Decoder. For the Critic network, the hidden size is also set to 128. 

\begin{table}[htbp]
  \renewcommand{\arraystretch}{1.3}
  \centering
  \caption{Parameter settings of the model. 1D-Conv means the 1-D convolution layer. $D_{input}$ represents the dimension of input. Kernel size and stride are essential parameters of the 1-D convolution layer}
    \label{parameter:model}
    \begin{tabular}{cc}
    \toprule
    \multicolumn{2}{c}{Actor network(Pointer Network)} \\
    
    \midrule
    \multicolumn{1}{l}{Encoder:} & \multicolumn{1}{l}{1D-Conv($D_{input}$, 128, kernel size=1, stride=1) } \\
    \multicolumn{1}{l}{Decoder:} & \multicolumn{1}{l}{GRU(hidden size=128, number of layer=1)} \\
          & \multicolumn{1}{l}{Attention(No hyper parameters)} \\
    \toprule
    \multicolumn{2}{c}{Critic network} \\
    
    \midrule
    \multicolumn{2}{c}{1D-Conv($D_{input}$, 128, kernel size=1, stride =1)} \\
    \multicolumn{2}{c}{1D-Conv(128, 20, kernel size=1, stride =1)} \\
    \multicolumn{2}{c}{1D-Conv(20, 20, kernel size=1, stride =1)} \\
    \multicolumn{2}{c}{1D-Conv(20, 1, kernel size=1, stride =1)} \\
    \bottomrule
    \end{tabular}%
 
\end{table}%

We train both of the actor and critic networks using the Adam optimizer \cite{kingma2014adam} with learning rate $\eta$ of 0.0001 and batch size of 200. The \textit{Xavier} initialization method \cite{glorot2010understanding} is used to initialize the weights for the first subproblem. Weights for the following subproblems are generated by the introduced neighborhood-based parameter transfer strategy. 

In addition, different size of generated instances are required for training different types of models. As compared with the Mixed MOTSP problem, the model of Euclidean MOTSP problem requires more weights to be optimized because its dimension of input is larger, thus requiring more training instances in each iteration. In this work, we generate 500,000 instances to train the Euclidean bi-objective TSP and 120,000 instances to train the Mixed one. All the problem instances are generated from an  uniform distribution of $[0,1]$ and used for training for 5 epochs. It costs about 3 hours to train the Mixed instances and 7 hours to train the Euclidean instances. Once the model is trained, it can be used to directly output the Pareto Fronts. 

\section{Experimental Results and Discussions}
\label{result}
In this section, DRL-MOA is compared with classical MOEAs of NSGA-II and MOEA/D on different MOTSP instances. The maximum number of iteration for NSGA-II and MOEA/D is set to 500, 1000, 2000 and 4000 respectively. The population size is set to 100 for NSGA-II and MOEA/D. The number of subproblems for DRL-MOA is set to 100 as well. The Tchebycheff approach, which we found would perform better on MOTSP, is used for MOEA/D. In addition, only the non-dominated solutions are reserved in the final PF.
\subsection{Results on Mixed type bi-objective TSP}

\par We first test the model that is trained on 40-city Mixed type bi-objective TSP instances. The model is then used to approximate the PF of 40-, 70-, 100-, 150- and 200-city instances. 

\begin{figure}[htbp]
\centering
\subfloat[DRL-MOA and NSGA-II]{\includegraphics[width=1.5in]{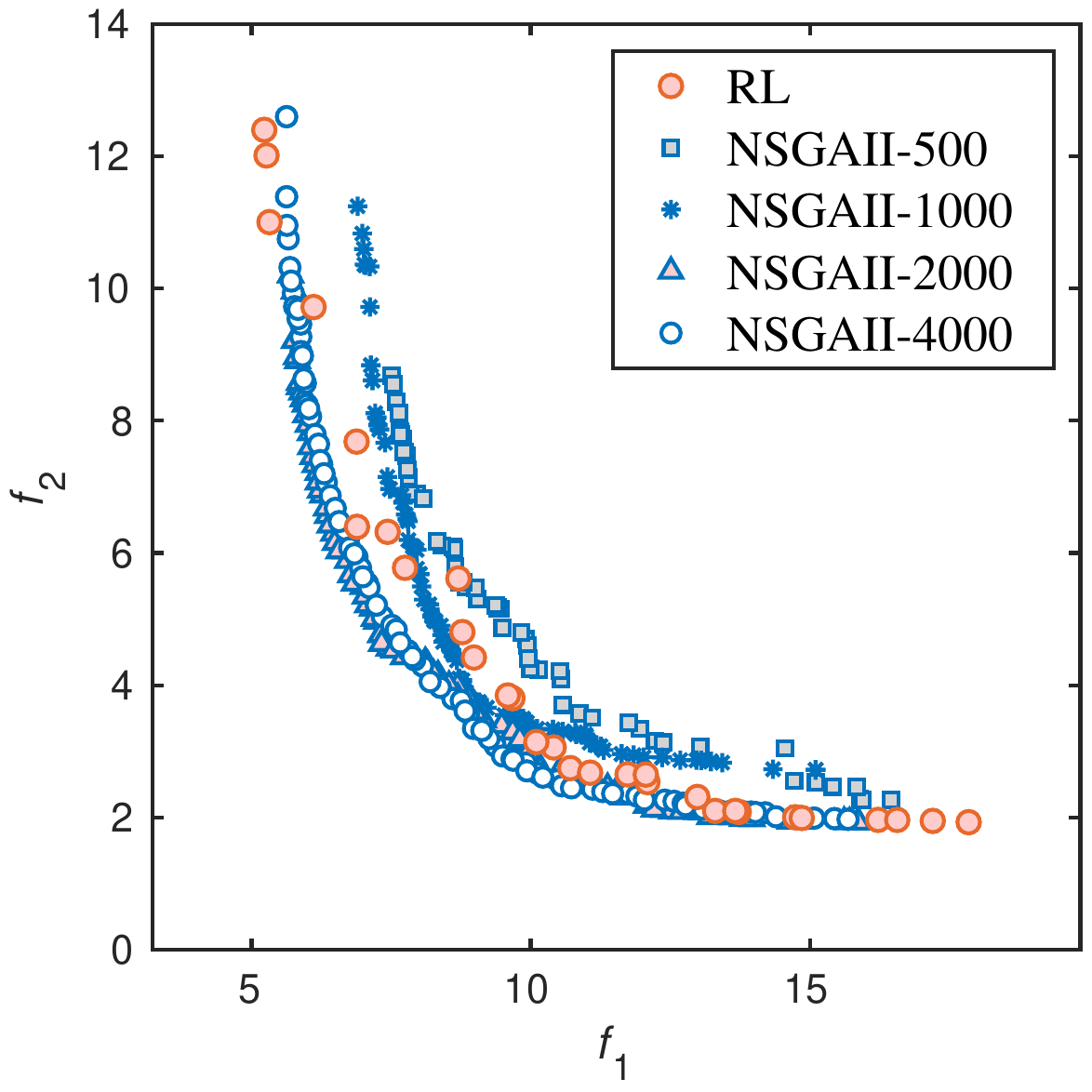}%
}
\hfil
\subfloat[DRL-MOA and MOEA/D]{\includegraphics[width=1.5in]{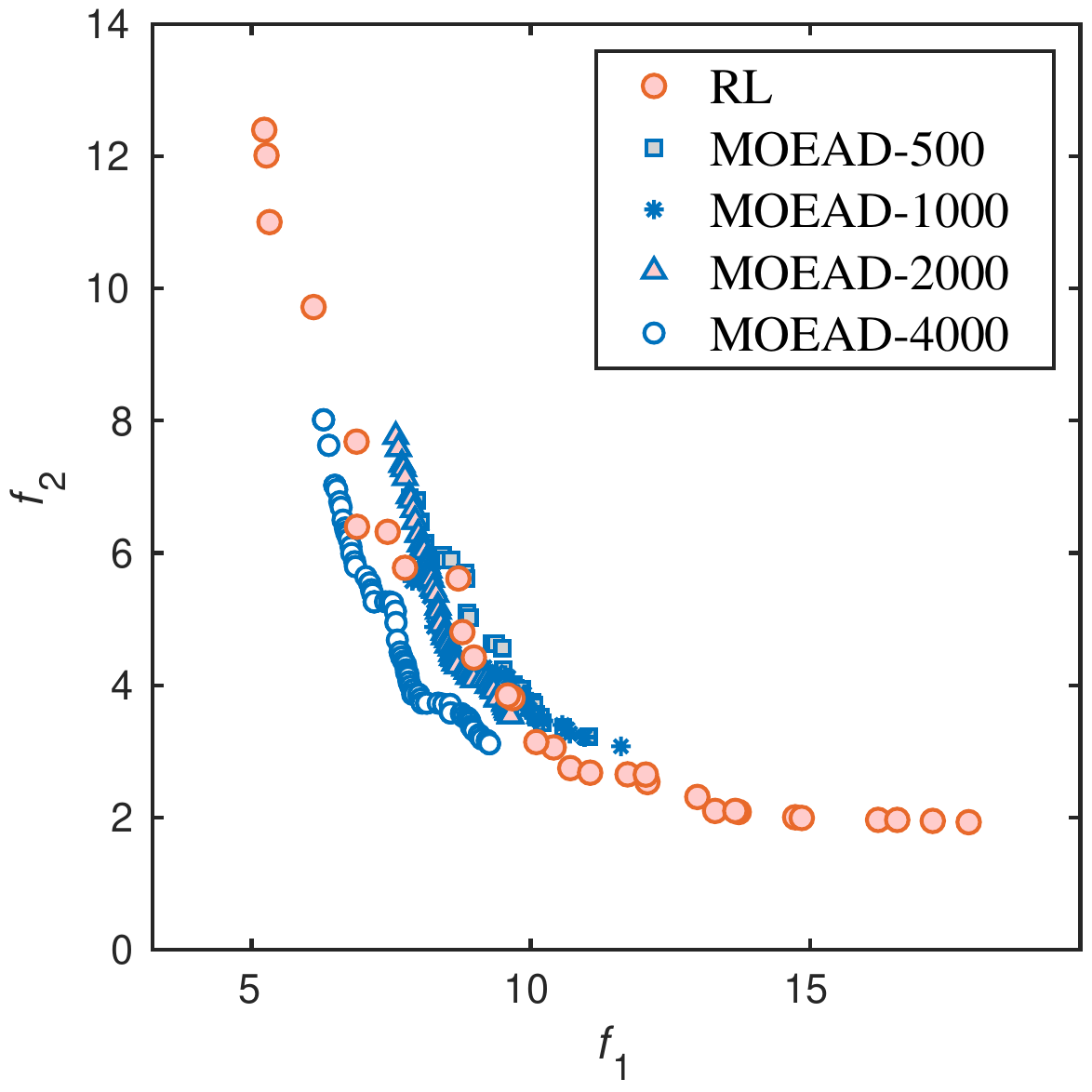}%
}
\caption{A random generated \textbf{40-city} Mixed bi-objective TSP problem instance: the PF obtained using our method (trained using 40-city instances) in comparison with NSGA-II and MOEA/D. 500, 1000, 2000, 4000 iterations are applied respectively.}
\label{40city}
\end{figure}

\begin{figure}[htbp]
\centering
\subfloat[DRL-MOA and NSGA-II]{\includegraphics[width=1.5in]{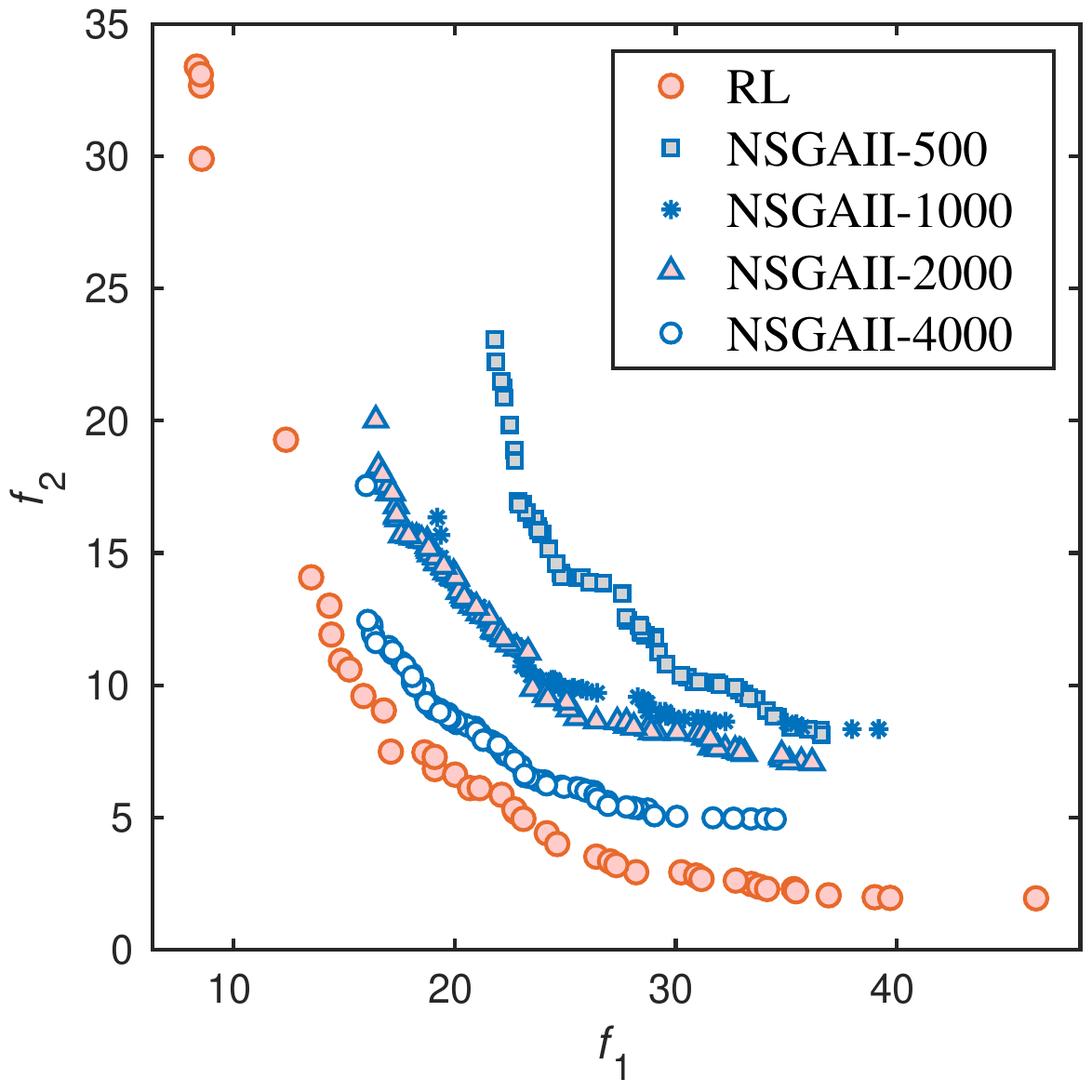}%
}
\hfil
\subfloat[DRL-MOA and MOEA/D]{\includegraphics[width=1.5in]{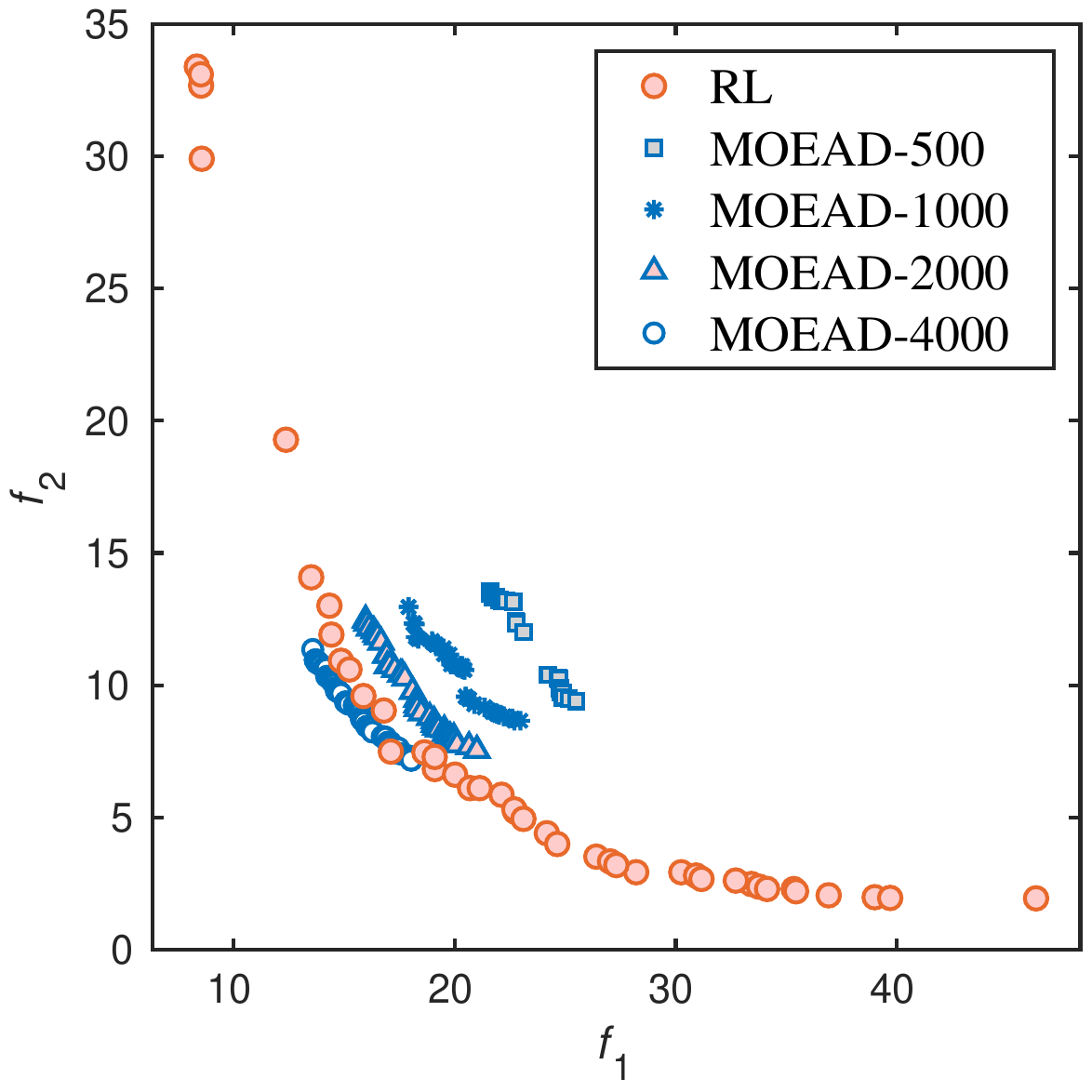}%
}
\caption{A random generated \textbf{100-city} Mixed bi-objective TSP problem instance: the PF obtained using our method (trained using 40-city instances) in comparison with NSGA-II and MOEA/D. 500, 1000, 2000, 4000 iterations are applied respectively.}
\label{100city}
\end{figure}

\begin{figure}[htbp]
\centering
\subfloat[DRL-MOA and NSGA-II]{\includegraphics[width=1.5in]{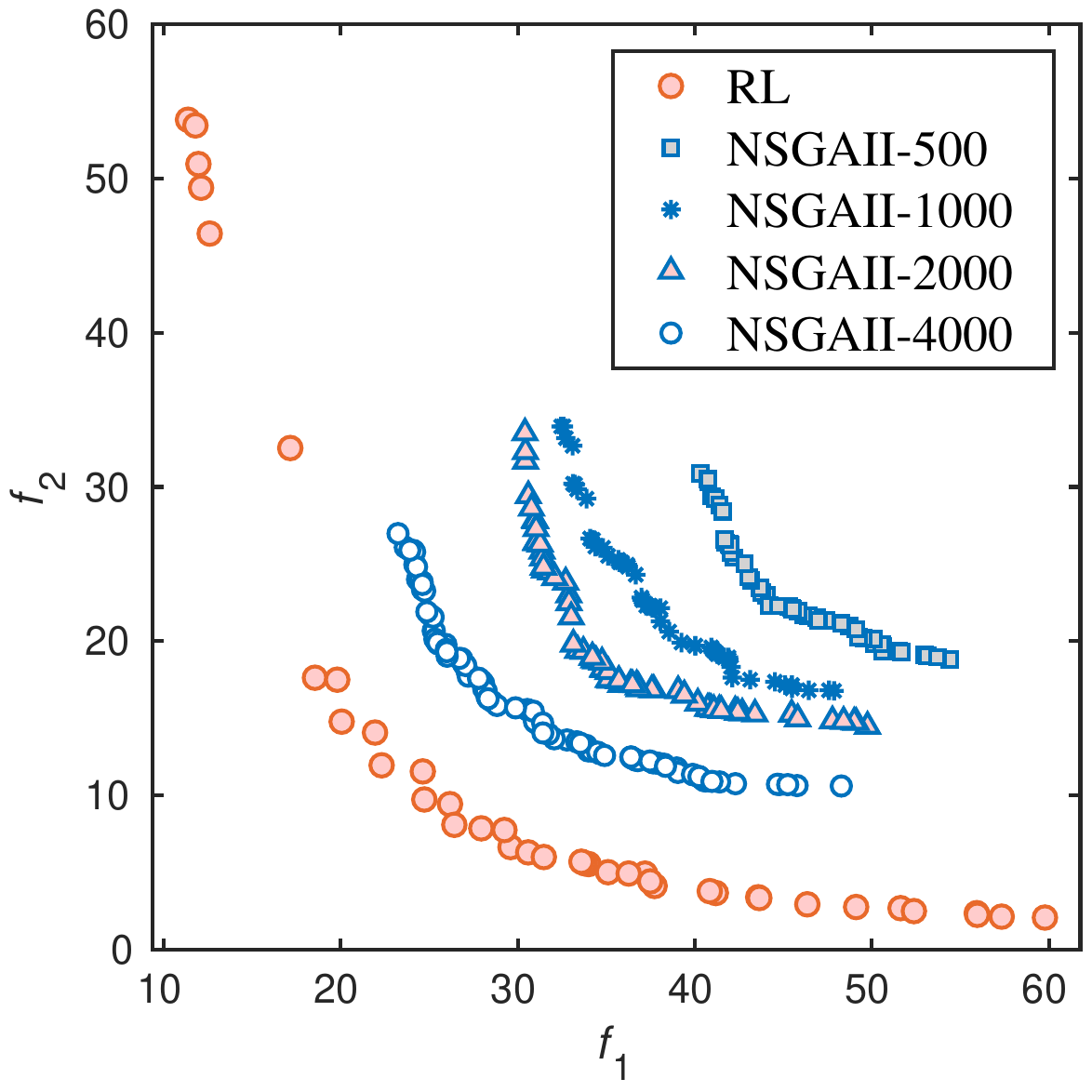}%
}
\hfil
\subfloat[DRL-MOA and MOEA/D]{\includegraphics[width=1.5in]{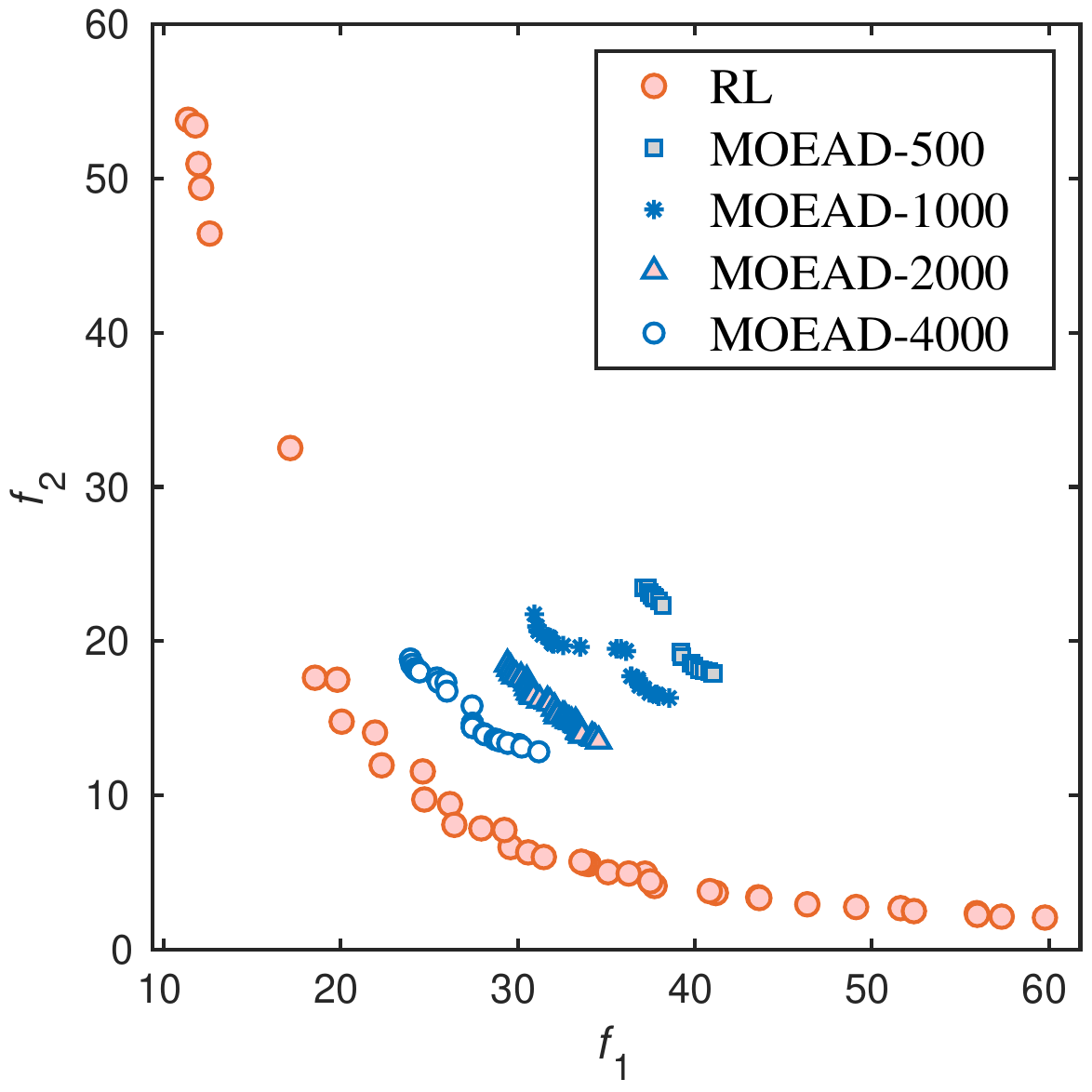}%
}
\caption{A random generated \textbf{150-city} Mixed bi-objective TSP problem instance: the PF obtained using our method (trained using 40-city instances) in comparison with NSGA-II and MOEA/D. 500, 1000, 2000, 4000 iterations are applied respectively.}
\label{150city}
\end{figure}

\begin{figure}[htbp]
\centering
\subfloat[DRL-MOA and NSGA-II]{\includegraphics[width=1.5in]{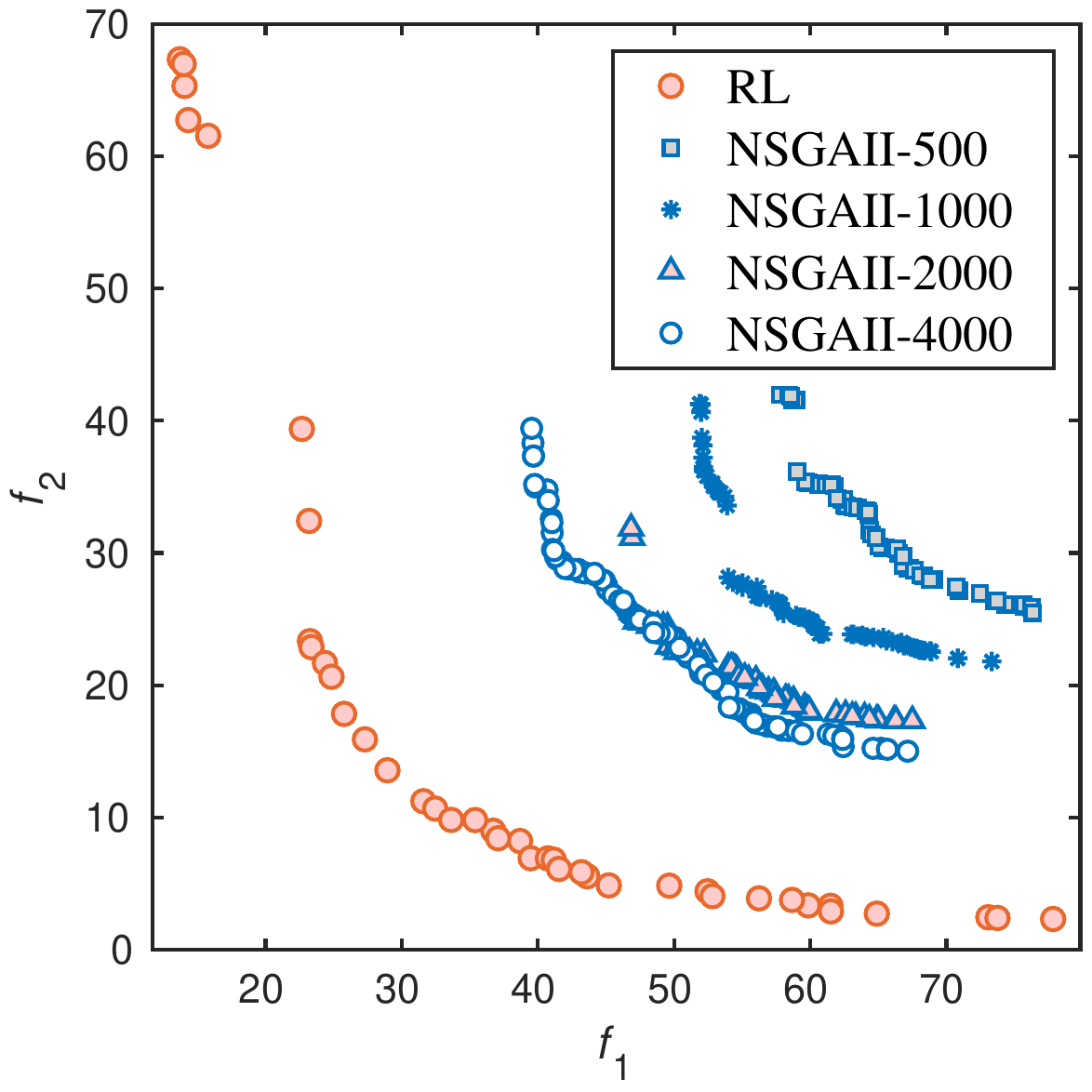}%
}
\hfil
\subfloat[DRL-MOA and MOEA/D]{\includegraphics[width=1.5in]{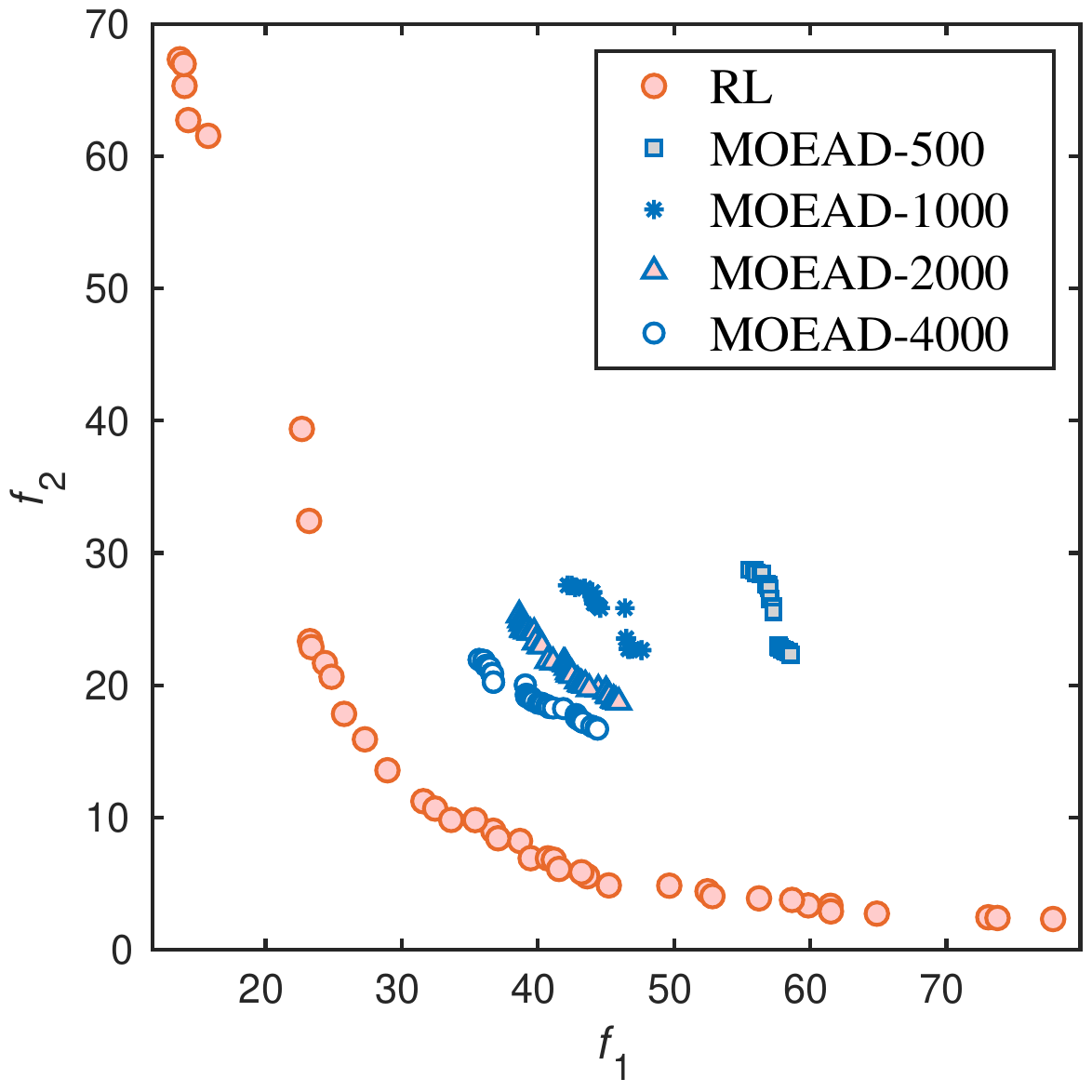}%
}
\caption{A random generated \textbf{200-city} Mixed bi-objective TSP problem instance: the PF obtained using our method (trained using 40-city instances) in comparison with NSGA-II and MOEA/D. 500, 1000, 2000, 4000 iterations are applied respectively.}
\label{200city}
\end{figure}

The performance of the PFs obtained by all the compared algorithms on various instances are shown in Fig. \ref{40city}, \ref{100city}, \ref{150city}, \ref{200city}. It is observed that the trained model can efficiently scale to bi-objective TSP with different number of cities. Although the model is obtained by training on the 40-city instances, it can still exhibit good performances on the 70-, 100-, 150- and 200-city instances. Moreover, the performance indicator of Hypervolume (HV) and the running time that are obtained based on five runs are also listed in Table \ref{tab:mixed}. 

As shown in \fref{40city}, all of the compared algorithms can work well for the small-scale problems, e.g., 40-city instances. By increasing the number of iterations, NSGA-II and MOEA/D even show a better ability of convergence. However, the large number of iterations can lead to a large amount of computing time. For example, 4000 iterations cost 130.2 seconds for MOEA/D and 28.3 seconds for NSGA-II while our method just requires 2.7 seconds. 

As can be seen in Fig. \ref{100city}, \ref{150city}, \ref{200city}, as the number of cities increases, the competitors of NSGA-II and MOEA/D struggle to converge while the DRL-MOA exhibits a much better ability of convergence. 

For 100-city instances in \fref{100city}, MOEA/D shows a slightly better performance in terms of convergence than other methods by running 4000 iterations with 140.3 seconds. However, the diversity of solutions found by our method is much better than MOEA/D. 

For 150- and 200-city instances as depicted in \fref{150city} and \fref{200city}, NSGA-II and MOEA/D exhibit an obviously inferior performance than our method in terms of both the convergence and diversity. Even though the competitors are conducted for 4000 iterations, which is a pretty large number of iterations, DRL-MOA still shows a far better performance than them.

In addition, the DRL-MOA achieves the best HV comparing to other algorithms, as shown in \tref{tab:mixed}. Also, its running time is much lower in comparison with the competitors. Overall, the experimental results clearly indicate the effectiveness of DRL-MOA on solving large-scale bi-objective TSPs. The \emph{brain} of the trained model has learned how to select the next city given the city information and the selected cities. Thus it does not suffer the deterioration of performance with the increasing number of cities. In contrast, NSGA-II and MOEA/D fail to converge within a reasonable computing time for large-scale bi-objective TSPs. In addition, the PF obtained by the DRL-MOA method shows a significantly better diversity as compared with NSGA-II and MOEA/D whose PF have a much smaller spread.

\begin{table*}[!t]
  \centering
  \caption{HV values obtained by DRL-MOA, NSGA-II and MOEA/D. Instances of 40-, 70-, 100-, 150-, 200-city \textbf{Mixed} type bi-objective TSP are test. The running time is listed. The best HV is marked in gray background and the longest running time is marked bold}
    \begin{tabular}{ccccccccccc}
    \toprule
          & \multicolumn{2}{c}{40-city} & \multicolumn{2}{c}{70-city} & \multicolumn{2}{c}{100-city} & \multicolumn{2}{c}{150-city} & \multicolumn{2}{c}{200-city} \\
    \midrule
          & HV    & Time/s & HV    & Time/s & HV    & Time/s & HV    & Time/s & HV    & Time/s \\
    \midrule
    NSGAII-500 & 1282  & 4.1   & 3866  & 4.2   & 7186  & 4.6   & 15158  & 5.7   & 26246  & 6.5  \\
    NSGAII-1000 & 1345  & 7.0   & 4042  & 9.6   & 7717  & 8.7   & 16313  & 12.6  & 27557  & 11.9  \\
    NSGAII-2000 & 1366  & 13.3  & 4146  & 16.8  & 8218  & 16.3  & 17283  & 21.4  & 29206  & 23.3  \\
    NSGAII-4000 & \cellcolor[rgb]{ .749,  .749,  .749}1404  & 28.3  & 4434  & 32.7  & 8597  & 33.2  & 18267  & 40.5  & 31647  & 51.2  \\
    MOEA/D-500 & 1251  & 17.0  & 3878  & 17.7  & 7367  & 18.5  & 15796  & 20.5  & 26548  & 21.8  \\
    MOEA/D-1000 & 1305  & 34.5  & 4048  & 35.2  & 7796  & 35.9  & 16838  & 40.6  & 28851  & 41.9  \\
    MOEA/D-2000 & 1324  & 65.2  & 4166  & 68.5  & 8261  & 73.2  & 17833  & 79.4  & 30785  & 85.5  \\
    MOEA/D-4000 & 1346  & \textbf{130.2 } & 4235  & \textbf{136.0 } & 8471  & \textbf{145.2 } & 18644  & \textbf{157.6 } & 32642  & \textbf{169.2 } \\
    DRL-MOA & 1398  & 2.7     & \cellcolor[rgb]{ .749,  .749,  .749}4668  & 4.7     & \cellcolor[rgb]{ .749,  .749,  .749}9647  & 6.6    & \cellcolor[rgb]{ .749,  .749,  .749}22386  & 10.1    & \cellcolor[rgb]{ .749,  .749,  .749}40354  & 12.9 \\
    \bottomrule
    \end{tabular}%
  \label{tab:mixed}%
\end{table*}%

\subsection{Results on Euclidean type bi-objective TSP}

\par We then test the model on Euclidean type instances. The DRL-MOA model is trained on 40-city instances and applied to approximate the PF of 40-, 70-, 100-, 150- and 200-city instances. For 100-, 150- and 200-city problems, we adopt the commonly used kroAB100, kroAB150 and kroAB200 instances \cite{lust2010multiobjective}. The HV indicator and computing time are shown in  \tref{tab:euclidean}.

\begin{table*}[!t]
  \centering
  \caption{HV values obtained by DRL-MOA, NSGA-II and MOEA/D. Instances of 40-, 70-, 100-, 150-, 200-city \textbf{Euclidean} type bi-objective TSP are test. The running time is listed. The best HV is marked in gray background and the longest running time is marked bold}
    \begin{tabular}{ccccccccccc}
    \toprule
          & \multicolumn{2}{c}{40-city} & \multicolumn{2}{c}{70-city} & \multicolumn{2}{c}{100-city} & \multicolumn{2}{c}{150-city} & \multicolumn{2}{c}{200-city} \\
    \midrule
          & HV    & Time/s & HV    & Time/s & HV    & Time/s & HV    & Time/s & HV    & Time/s \\
    \midrule
    NSGAII-500 & 1498  & 3.8   & 4446  & 4.1   & 8738  & 4.3   & 18487  & 5.2   & 31430  & 5.9  \\
    NSGAII-1000 & 1547  & 7.2   & 4643  & 7.7   & 9119  & 8.1   & 19491  & 9.6   & 33424  & 11.0  \\
    NSGAII-2000 & 1587  & 13.4  & 4798  & 15.7  & 9577  & 15.6  & 20116  & 20.1  & 35261  & 23.3  \\
    NSGAII-4000 & 1596  & 26.7  & 4874  & 29.5  & 9816  & 31.8  & 21395  & 40.2  & 36375  & 54.1  \\
    MOEA/D-500 & 1485  & 16.4  & 4438  & 17.1  & 8851  & 18.5  & 18941  & 20.3  & 32540  & 21.2  \\
    MOEA/D-1000 & 1494  & 33.6  & 4576  & 34.3  & 9256  & 36.5  & 19897  & 39.7  & 34842  & 42.4  \\
    MOEA/D-2000 & 1525  & 65.2  & 4703  & 69.5  & 9594  & 71.7  & 20723  & 78.6  & 36253  & 84.8  \\
    MOEA/D-4000 & 1512  & \textbf{130.3}  & 4781  & \textbf{135.2}  & 9778  & \textbf{141.7}  & 21522  & \textbf{156.9}  & 37687  & \textbf{168.2}  \\
    DRL-MOA & \cellcolor[rgb]{ .749,  .749,  .749}1603  & 2.6   & \cellcolor[rgb]{ .749,  .749,  .749}5150  & 4.5   & \cellcolor[rgb]{ .749,  .749,  .749}10773  & 6.3   & \cellcolor[rgb]{ .749,  .749,  .749}24567  & 9.4   & \cellcolor[rgb]{ .749,  .749,  .749}44110  & 12.9 \\
    \bottomrule
    \end{tabular}%
  \label{tab:euclidean}%
\end{table*}%

\begin{figure}[htbp]
\centering
\subfloat[DRL-MOA and NSGA-II]{\includegraphics[width=1.5in]{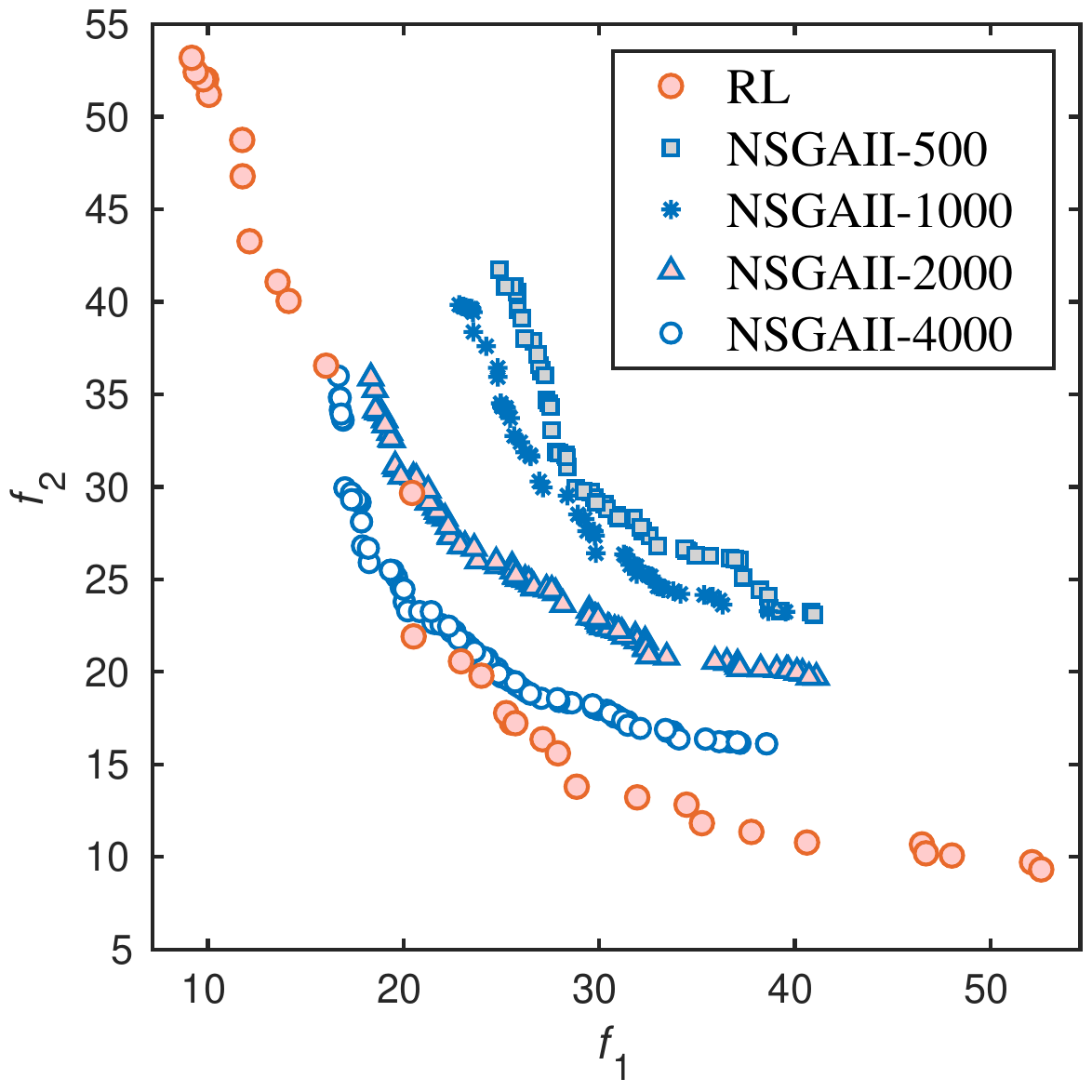}%
}
\hfil
\subfloat[DRL-MOA and MOEA/D]{\includegraphics[width=1.5in]{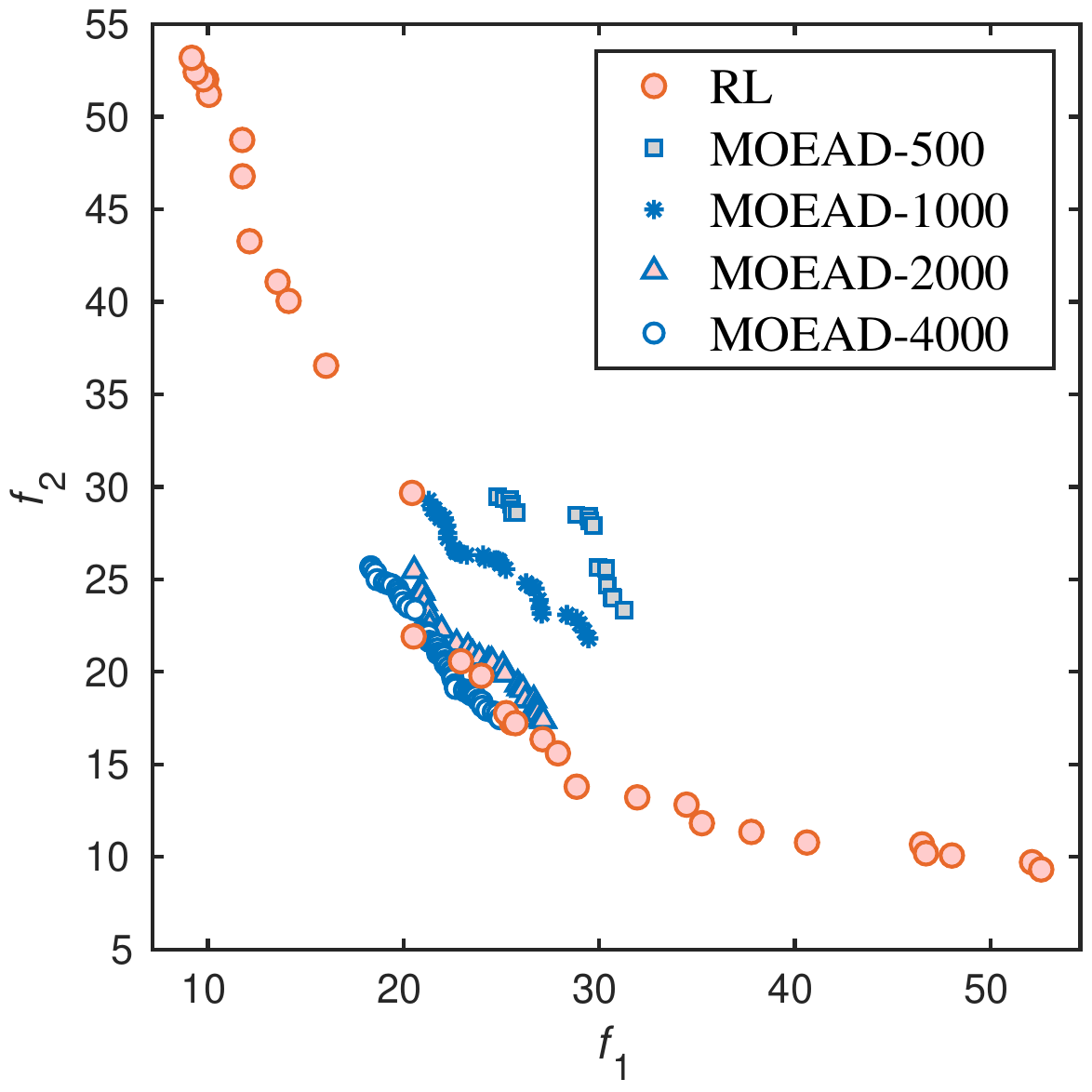}%
}
\caption{\textbf{KroAB100} Euclidean bi-objective TSP problem instance: the PF obtained using our method (trained using 40-city instances) in comparison with NSGA-II and MOEA/D. 500, 1000, 2000, 4000 iterations are applied respectively.}
\label{100city_4}
\end{figure}

\begin{figure}[htbp]
\centering
\subfloat[DRL-MOA and NSGA-II]{\includegraphics[width=1.5in]{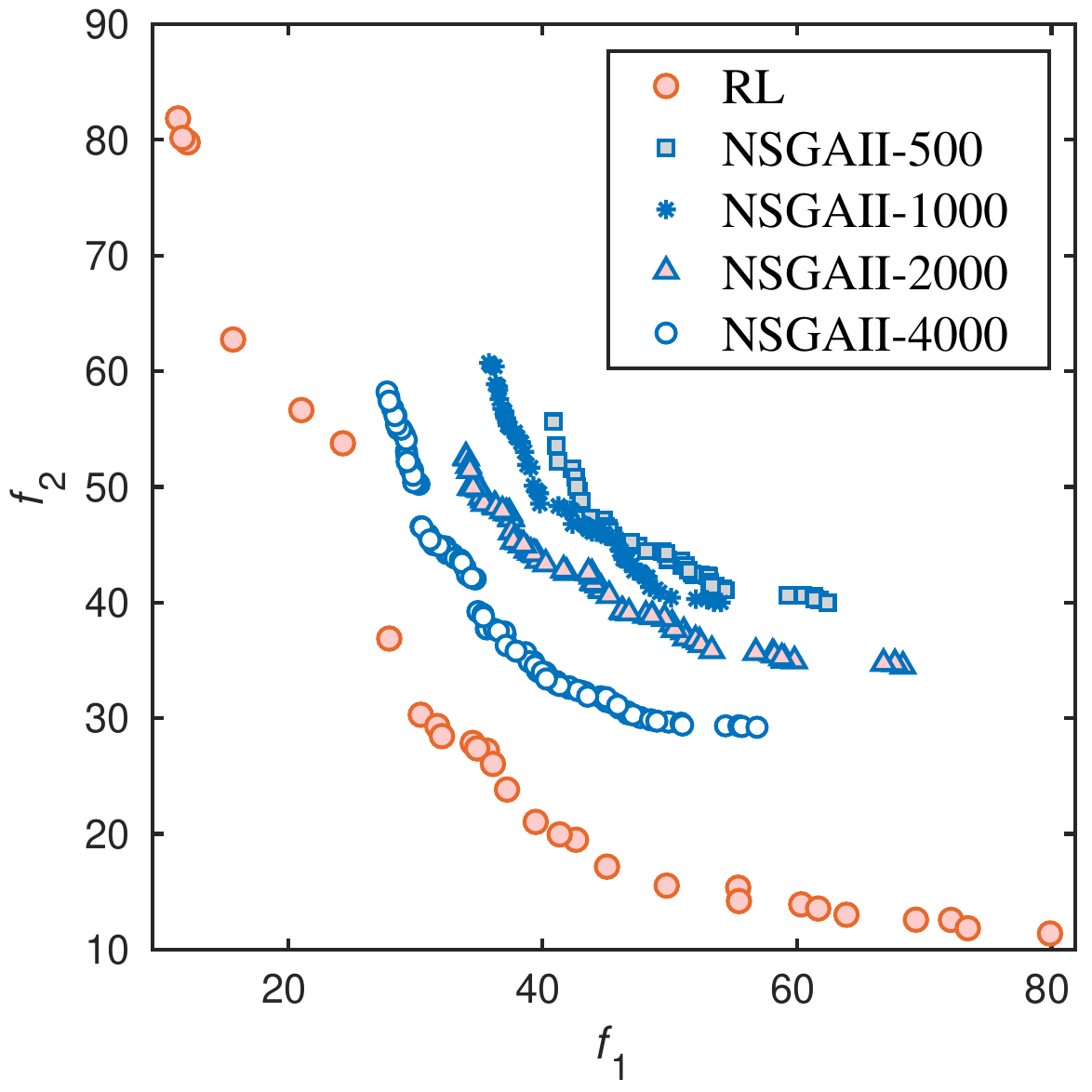}%
}
\hfil
\subfloat[DRL-MOA and MOEA/D]{\includegraphics[width=1.5in]{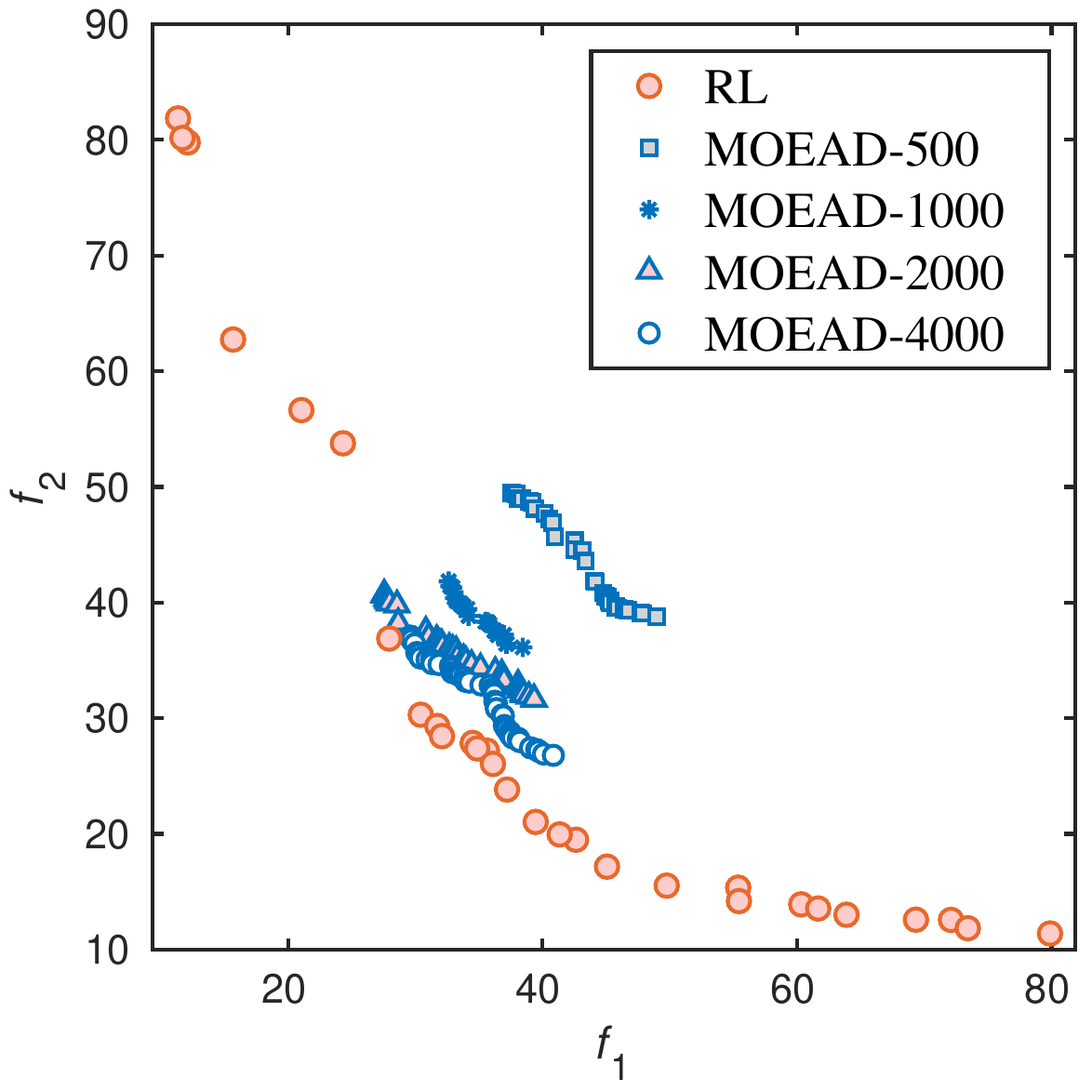}%
}
\caption{\textbf{KroAB150} Euclidean bi-objective TSP problem instance: the PF obtained using our method (trained using 40-city instances) in comparison with NSGA-II and MOEA/D. 500, 1000, 2000, 4000 iterations are applied respectively.}
\label{150city_4}
\end{figure}

\begin{figure}[htbp]
\centering
\subfloat[DRL-MOA and NSGA-II]{\includegraphics[width=1.5in]{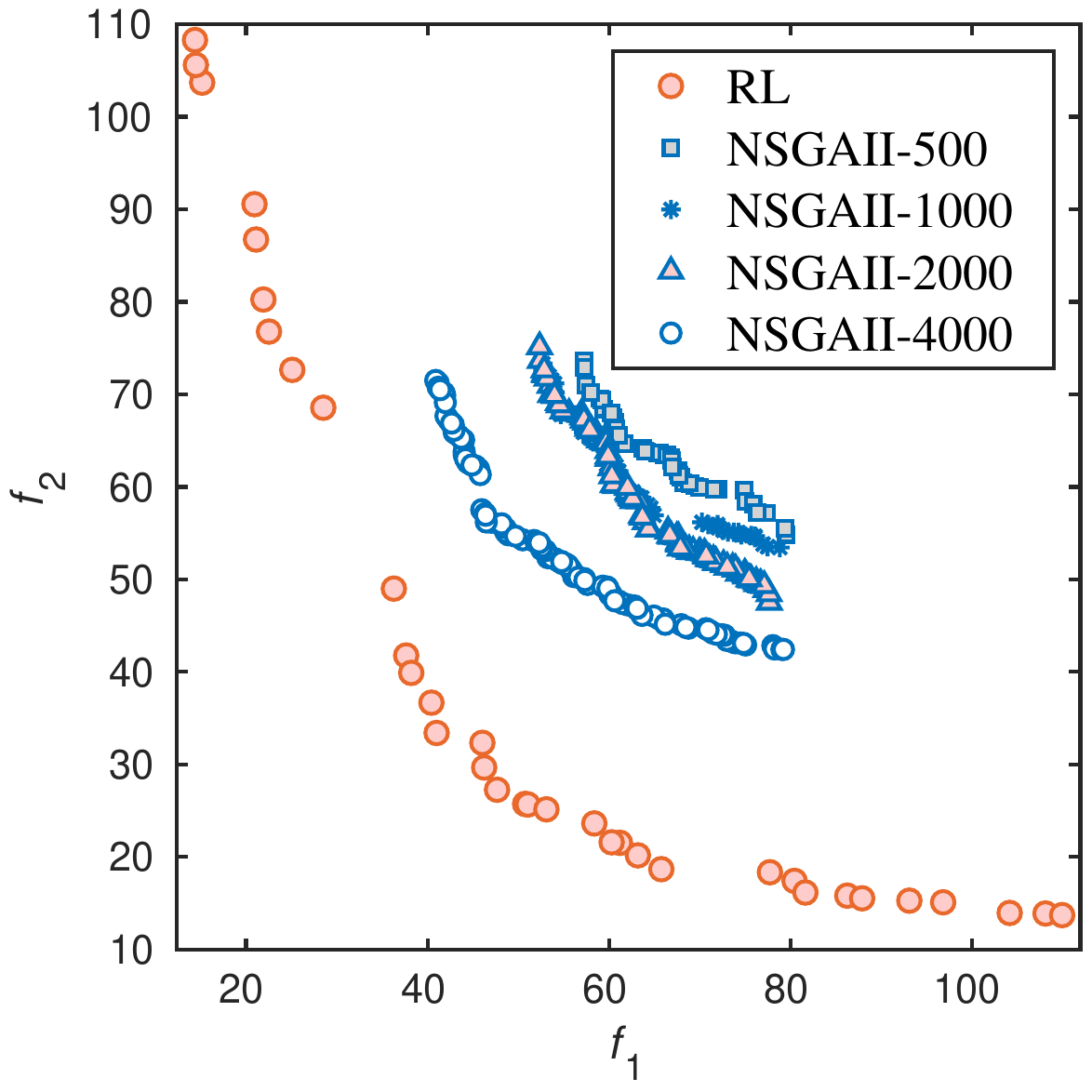}%
}
\hfil
\subfloat[DRL-MOA and MOEA/D]{\includegraphics[width=1.5in]{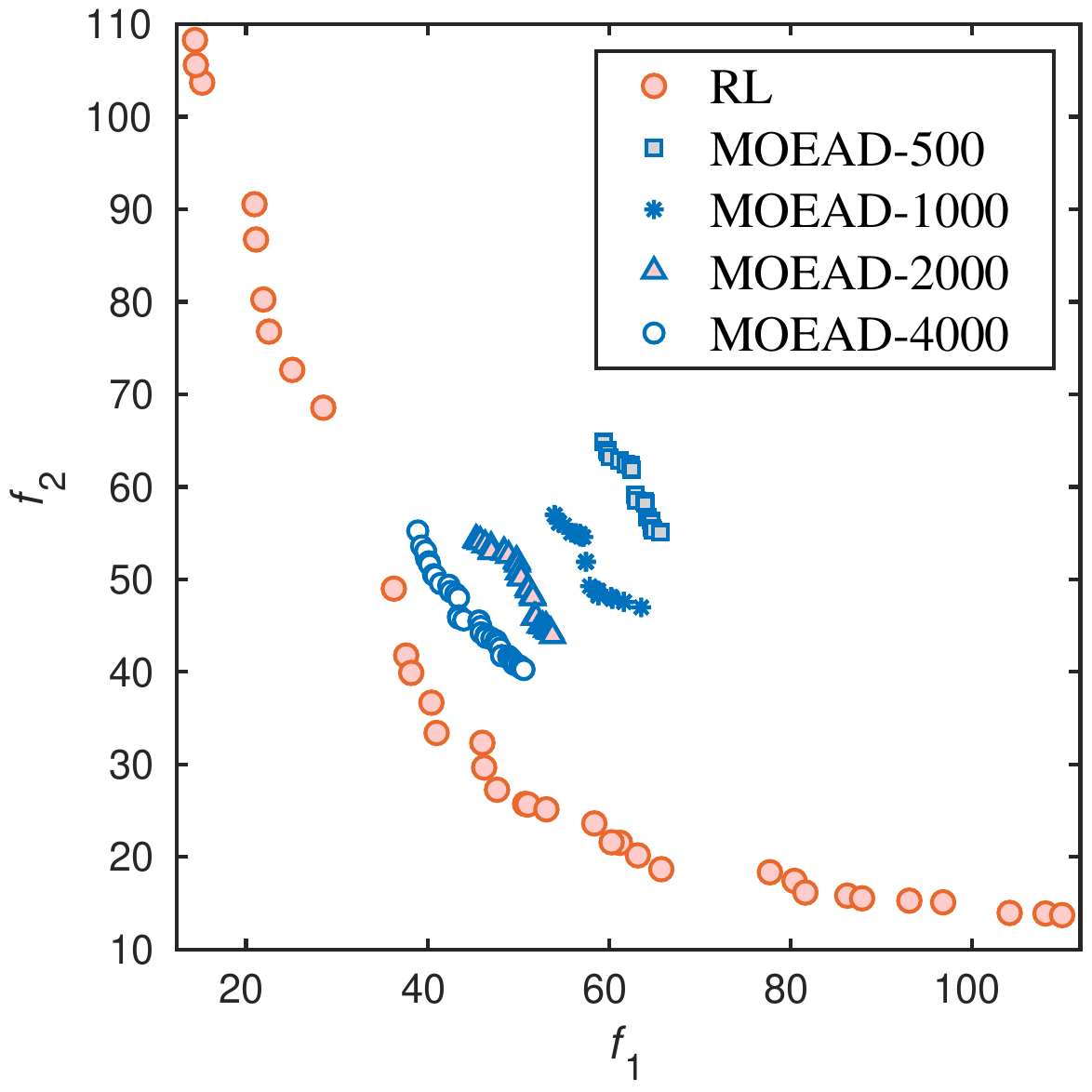}%
}
\caption{\textbf{KroAB200} Euclidean bi-objective TSP problem instance: the PF obtained using our method (trained using 40-city instances) in comparison with NSGA-II and MOEA/D. 500, 1000, 2000, 4000 iterations are applied respectively.}
\label{200city_4}
\end{figure}

Fig. \ref{100city_4}, \ref{150city_4} and \ref{200city_4} show the experimental results on kroAB100, kroAB150 and kroAB200 instances. For the kroAB100 instance, by increasing the number of iterations to 4000, NSGA-II, MOEA/D and DRL-MOA achieve a similar level of convergence while MOEA/D performs slightly better. However, MOEA/D performs the worst in terms of diversity with all solutions crowded in a small region and its running time is not acceptable. 

When the number of cities increases to 150 and 200, DRL-MOA significantly outperforms the competitors in terms of both convergence and diversity, as shown in \fref{150city_4} and \ref{200city_4}. Even though 4000 iterations are conducted for NSGA-II and MOEA/D, there is still an obvious gap of performance between the two methods and the DRL-MOA. 

\begin{figure}[htbp]
	\centering
	\subfloat[DRL-MOA and NSGA-II]{\includegraphics[width=1.5in]{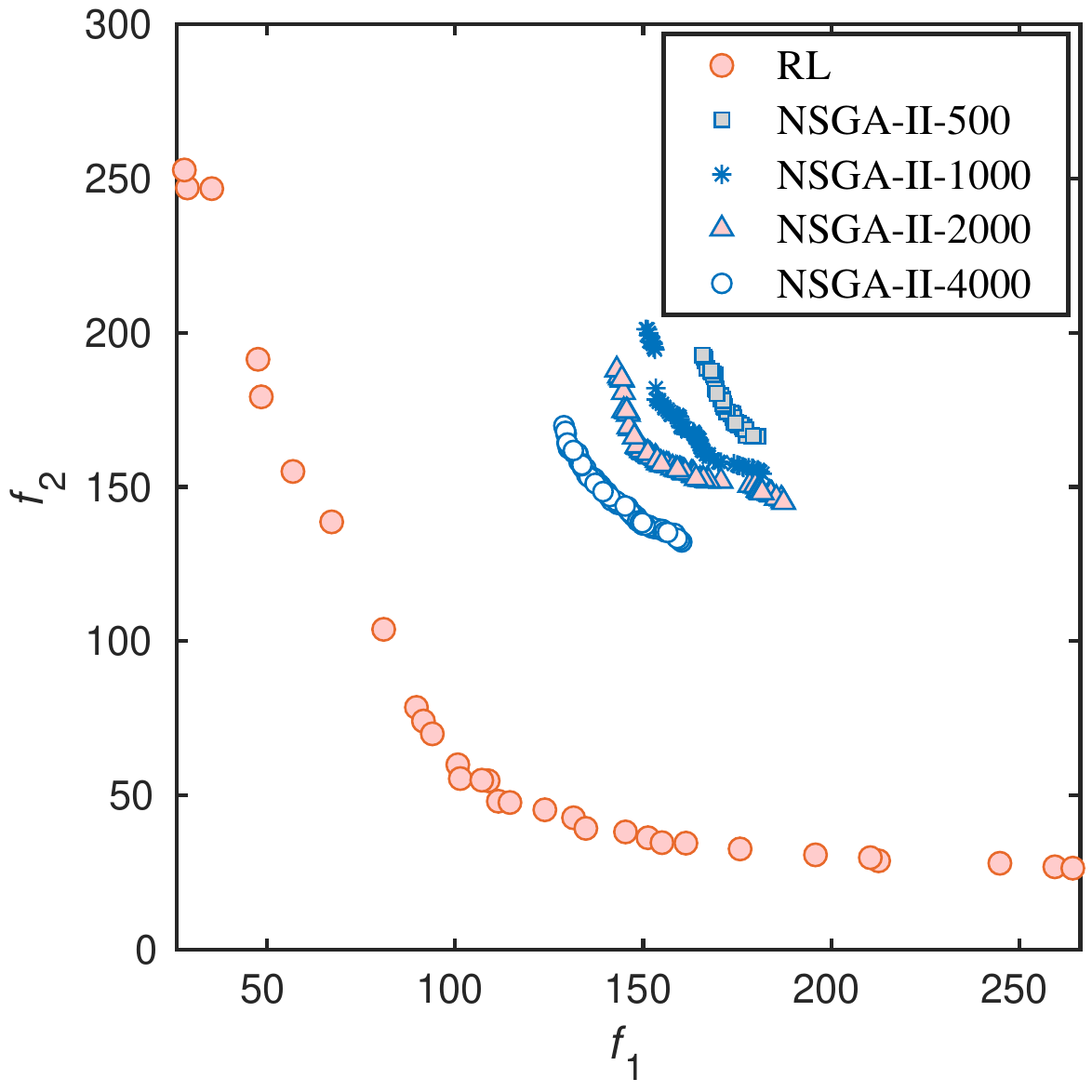}%
	}
	\hfil
	\subfloat[DRL-MOA and MOEA/D]{\includegraphics[width=1.5in]{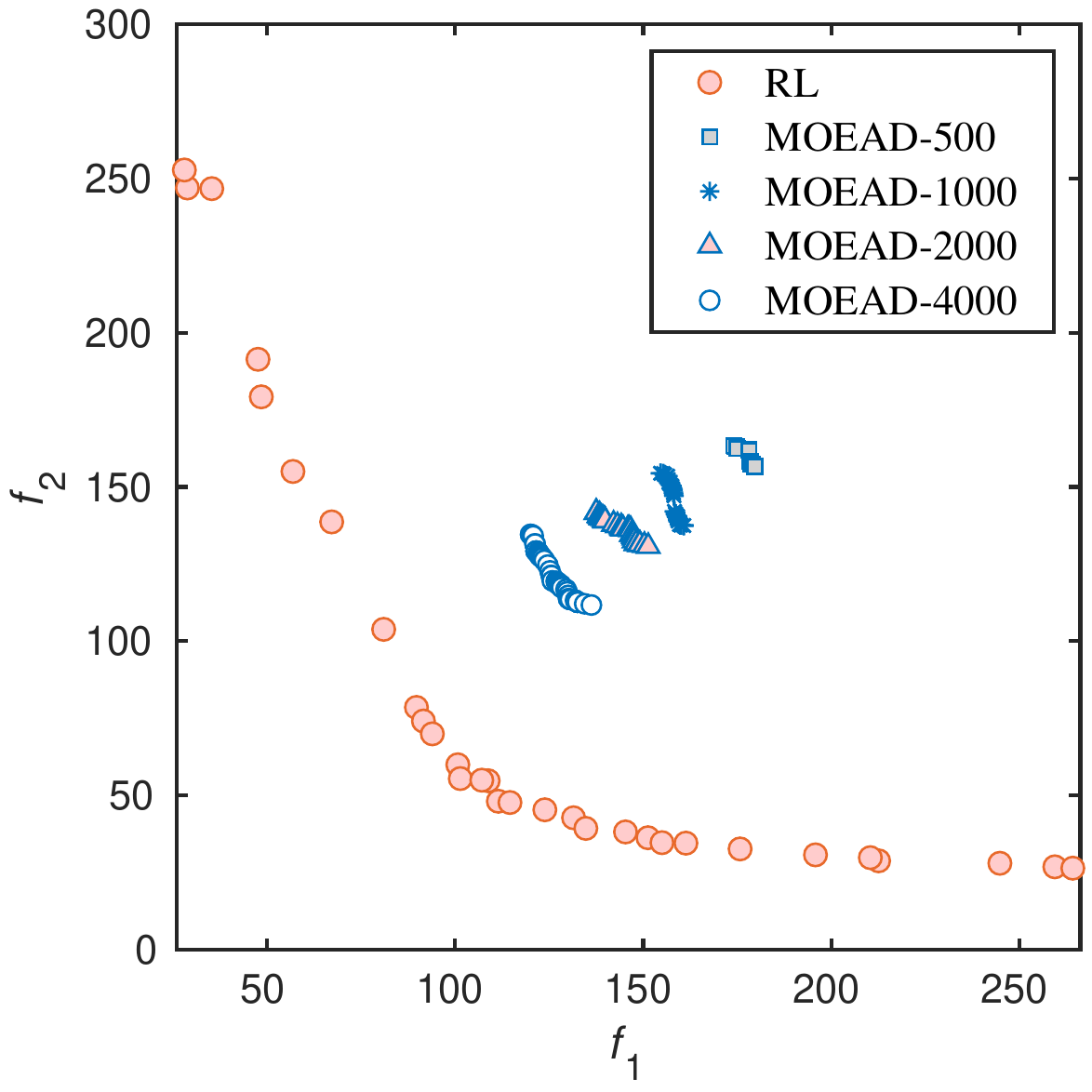}%
	}
	\caption{A random generated \textbf{500-city} Euclidean bi-objective TSP problem instance: the PF obtained using our method (trained using 40-city instances) in comparison with NSGA-II and MOEA/D. 500, 1000, 2000, 4000 iterations are applied respectively.}
	\label{500city_4}
\end{figure}

In terms of the HV indicator as demonstrated in \tref{tab:euclidean}, DRL-MOA performs the best on all instances. And the running time of DRL-MOA is much lower than the compared MOEAs. Increasing the number of iterations for MOEA/D and NSGA-II can certainly improve the performance but would result in a large amount of computing time. It requires more than 150 seconds for MOEA/D to reach an acceptable level of convergence. The computing time of NSGA-II is less, approximately 30 seconds, for running 4000 iterations. However, the performance for NSGA-II is always the worst amongst the compared methods. 

We further try to evaluate the performance of the model on 500-city instances. The model is still the one that is trained on 40-city instances and it is used to approximated the PF of a 500-city instance. The results are shown in \fref{500city_4}. It is observed that DRL-MOA significantly outperforms the competitors. And the performance gap is especially larger than that on smaller-scale problems.  

\subsection{Extension to MOTSPs with More Objectives}
In this section, the efficiency of our method is further evaluated on the 3- and 5-objective TSPs. The model is still trained on 40-city instances and it is used to approximate the PF of 100- and 200-city instances. The 3-objective TSP instances are constructed by combining two 2-dimensional inputs and a 1-dimensional input similar to the Mixed type instances. And the 5-objective TSP instances are constructed in the same way with two 2-dimensional inputs and three 1-dimensional input.

\begin{figure}[htbp]
	\centering
	\subfloat[DRL-MOA and NSGA-II]{\includegraphics[width=1.5in]{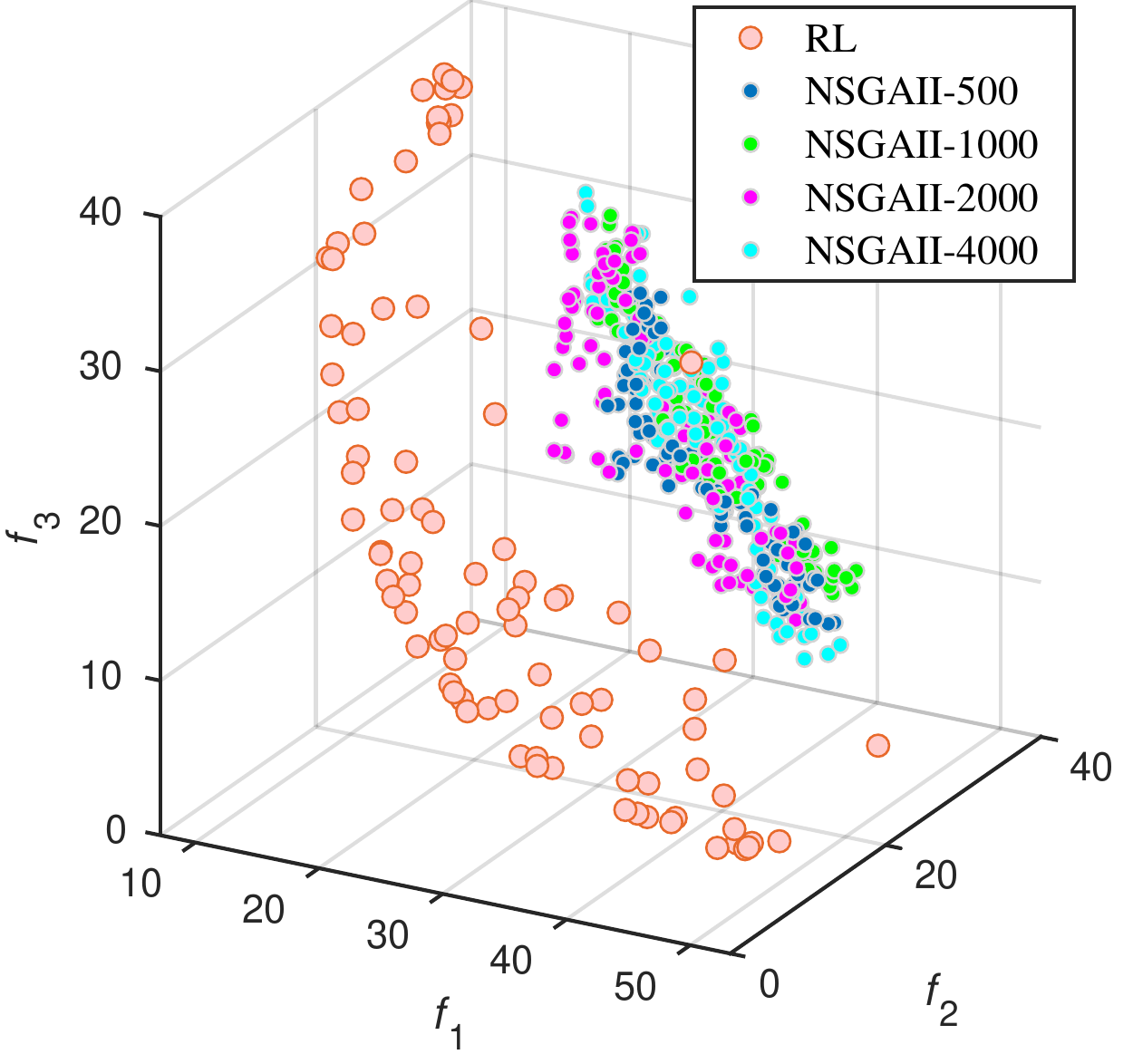}%
	}
	\hfil
	\subfloat[DRL-MOA and MOEA/D]{\includegraphics[width=1.5in]{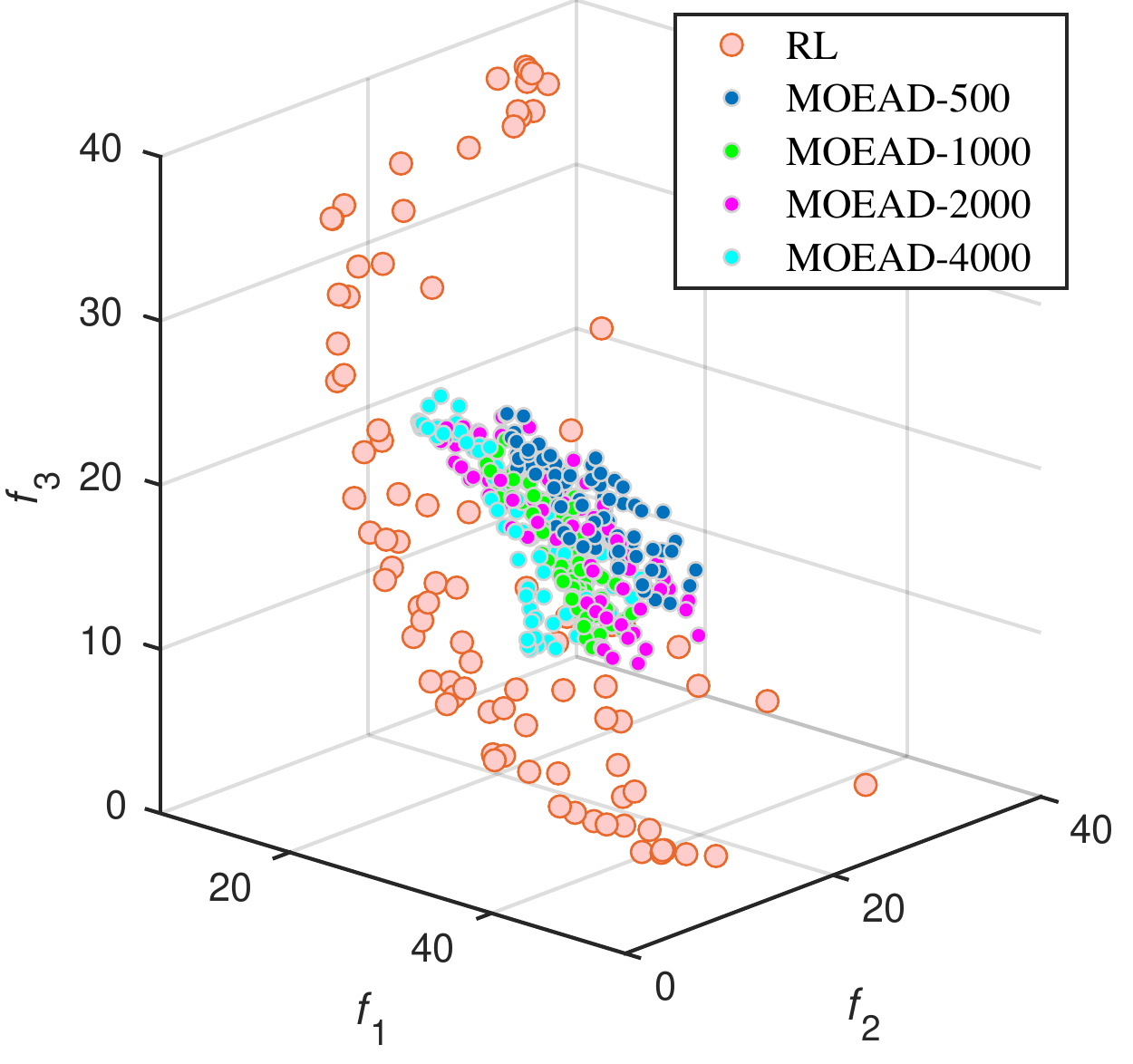}%
	}
	\caption{PFs obtained by DRL-MOA, NSGA-II and MOEA/D on a randomly generated 3-objective \textbf{100-city} TSP instance.}
	\label{100city_3obj}
\end{figure}

\begin{figure}[htbp]
	\centering
	\subfloat[DRL-MOA and NSGA-II]{\includegraphics[width=1.5in]{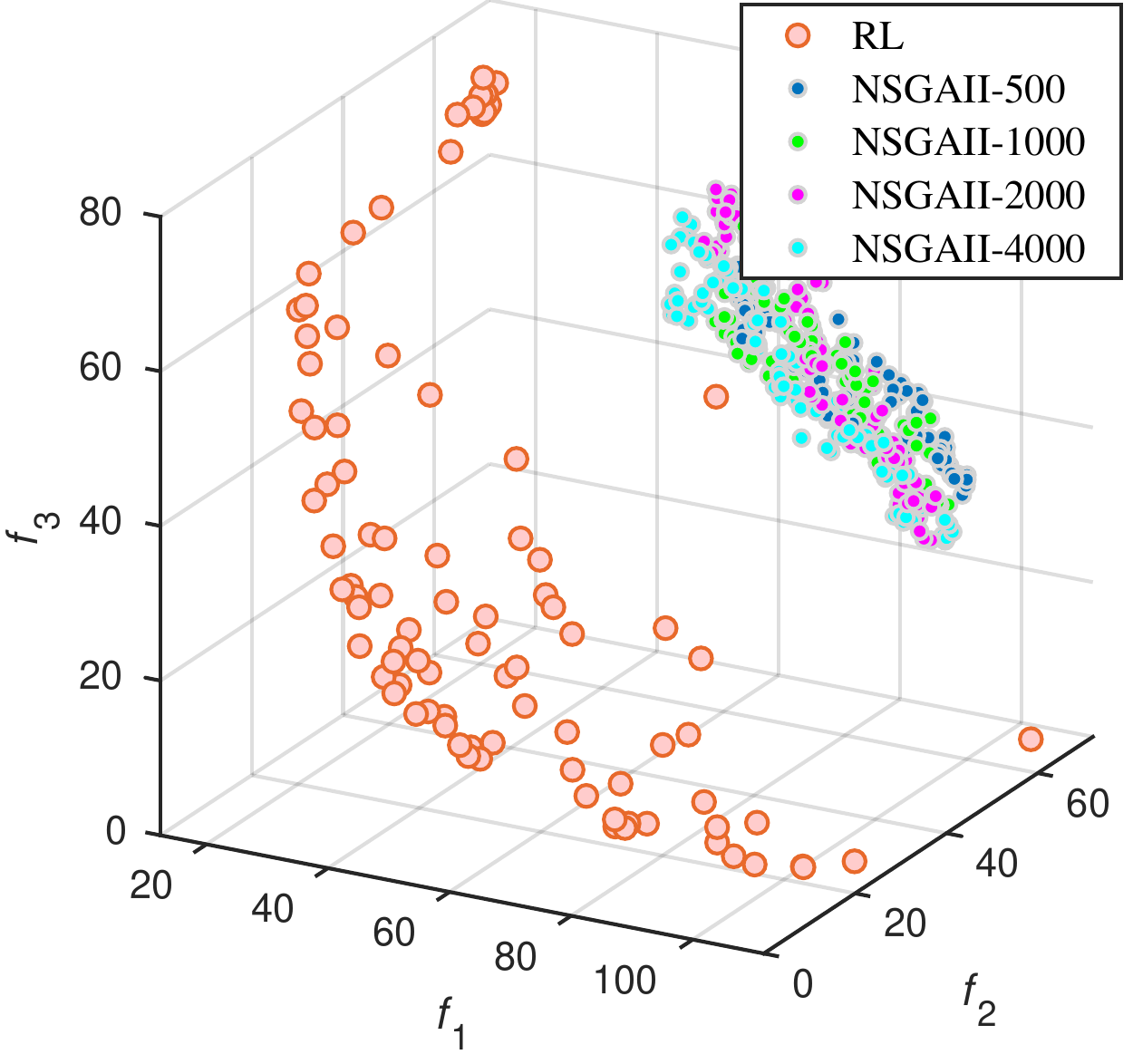}%
	}
	\hfil
	\subfloat[DRL-MOA and MOEA/D]{\includegraphics[width=1.5in]{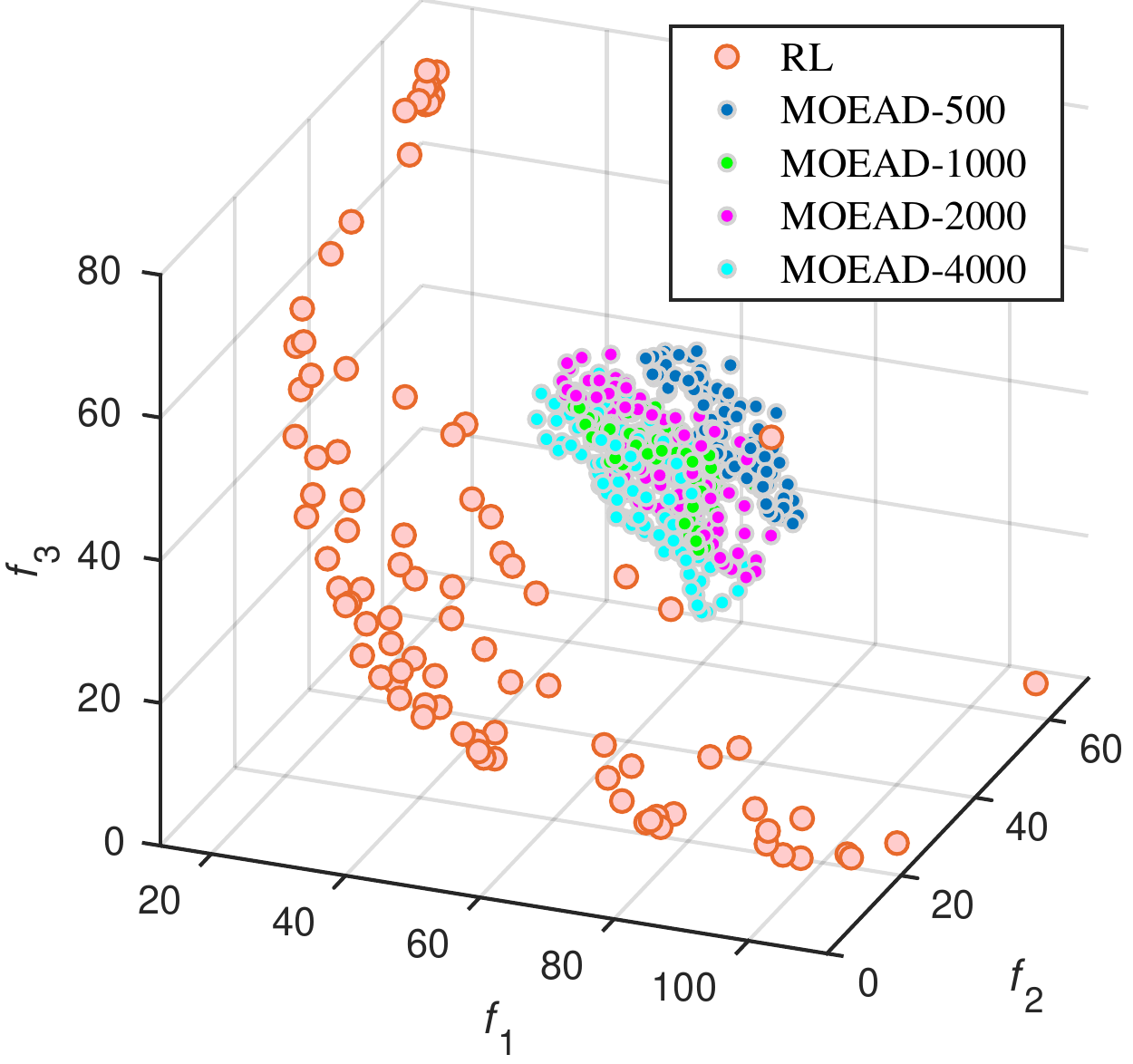}%
	}
	\caption{PFs obtained by DRL-MOA, NSGA-II and MOEA/D on a randomly generated 3-objective \textbf{200-city} TSP instance.}
	\label{200city_3obj}
\end{figure}

Results of the experiments on the 3-objective TSP are visualized in \fref{100city_3obj} and \fref{200city_3obj}. It can be observed that DRL-MOA significantly outperforms the classical MOEAs on all of the 100- and 200-city instances. Moreover, DRL-MOA performs clearly better on the 200-city instance than on the 100-city instance while the competitors struggle to converge.

In addition, results of the HV values on the 3- and 5-objective TSP instances that are obtained based on five runs are presented in \tref{5objs}. In can be seen that the DRL-MOA outperforms NSGA-II and MOEA/D on all instances. And NSGA-II is still the least effective method. 

\begin{table}[htbp]
  \centering
  \caption{HV values obtained by DRL-MOA, NSGA-II and MOEA/D on 3- and 5-objective TSP instances. The best HV is marked in gray background.}
    \begin{tabular}{crrrr}
    \toprule
          & \multicolumn{2}{c}{3-objective TSP} & \multicolumn{2}{c}{5-objective TSP} \\
    \midrule
          & \multicolumn{1}{l}{100-city} & \multicolumn{1}{l}{200-city} & \multicolumn{1}{l}{100-city} & \multicolumn{1}{l}{200-city} \\
    \midrule
    NSGAII-500 & 7.63E+05 & 5.09E+06 & 5.00E+09 & 1.31E+11 \\
    NSGAII-1000 & 7.72E+05 & 5.41E+06 & 5.19E+09 & 1.42E+11 \\
    NSGAII-2000 & 8.25E+05 & 5.56E+06 & 5.51E+09 & 1.57E+11 \\
    NSGAII-4000 & 8.53E+05 & 5.98E+06 & 6.12E+09 & 1.64E+11 \\
    MOEA/D-500 & 7.74E+05 & 5.63E+06 & 5.55E+09 & 1.53E+11 \\
    MOEA/D-1000 & 8.17E+05 & 6.02E+06 & 6.19E+09 & 1.55E+11 \\
    MOEA/D-2000 & 8.58E+05 & 6.26E+06 & 6.39E+09 & 1.68E+11 \\
    MOEA/D-4000 & 8.82E+05 & 6.49E+06 & 7.15E+09 & 1.78E+11 \\
    RL-MOA & \cellcolor[rgb]{ .651,  .651,  .651}1.13E+06 & \cellcolor[rgb]{ .651,  .651,  .651}9.42E+06 & \cellcolor[rgb]{ .651,  .651,  .651}1.19E+10 & \cellcolor[rgb]{ .651,  .651,  .651}4.06E+11 \\
    \bottomrule
    \end{tabular}%
  \label{5objs}%
\end{table}%

\subsection{Comparisons with Local-Search-based Methods}
In this section, DRL-MOA is compared with the local-search-based method. Ishibuchi and Murata first proposed a Multi-Objective Genetic Local Search Algorithm (MOGLS) \cite{ishibuchi1998multi} for multi-objective combinatorial optimization. It is further improved and specialized to solve the MOTSP in \cite{jaszkiewicz2002genetic}, which significantly outperforms the original MOGLS. In this section the DRL-MOA is compared with the improved MOGLS. 

Note that local search has been widely developed and various local-search-based methods that uses a number of specialized techniques to improve effectiveness have been proposed these years. However, the goal of our method is not to outperform a non-learned, specialized MOTSP algorithm. Rather, we show the fast solving speed and high generalization ability of our method via the combination of DRL and multi-objective optimization. Thus we did not consider more other local-search-based methods in this work. 

As no source code is found, the improved MOGLS is implemented by ourselves strictly according to \cite{jaszkiewicz2002genetic}. The algorithm is written in Python and run in the same machine to make fair comparisons and the code is publicly available\footnote{https://github.com/kevin031060/Genetic\_Local\_Search\_TSP}. 

The parameters, e.g., size of the temporary population and number of initial solution are consistent with the settings in \cite{jaszkiewicz2002genetic}. The local search uses a standard 2-opt algorithm, which is terminated if a pre-specified number of iterations $N_{LS}$ is completed. $N_{LS}$ is used to control the balance of computation time and model performance \cite{ishibuchi1998multi}. The improved MOGLS with $N_{LS}=100,200,300$ is used for comparisons. It should be noted that improving $N_{LS}$ or repeating the local search until no better solution is found might be able to further improve the performance of MOGLS. But it would be quite time-consuming and not experimented in this work. 

\begin{figure}[htbp]
	\centering
	\subfloat[PFs of Mixed instances]{\includegraphics[width=1.5in]{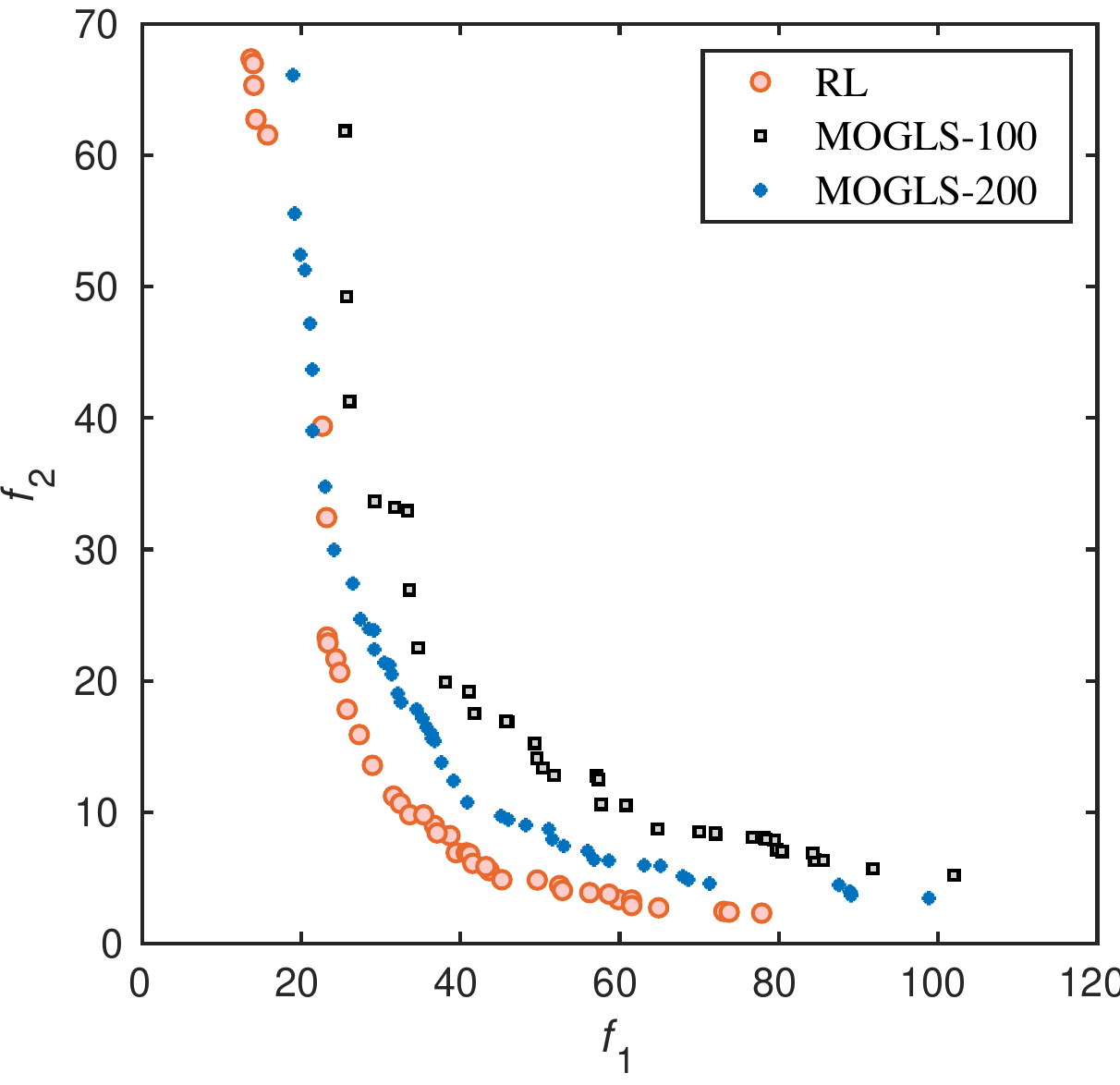}%
	}
	\hfil
	\subfloat[PFs of Euclidean instances ]{\includegraphics[width=1.5in]{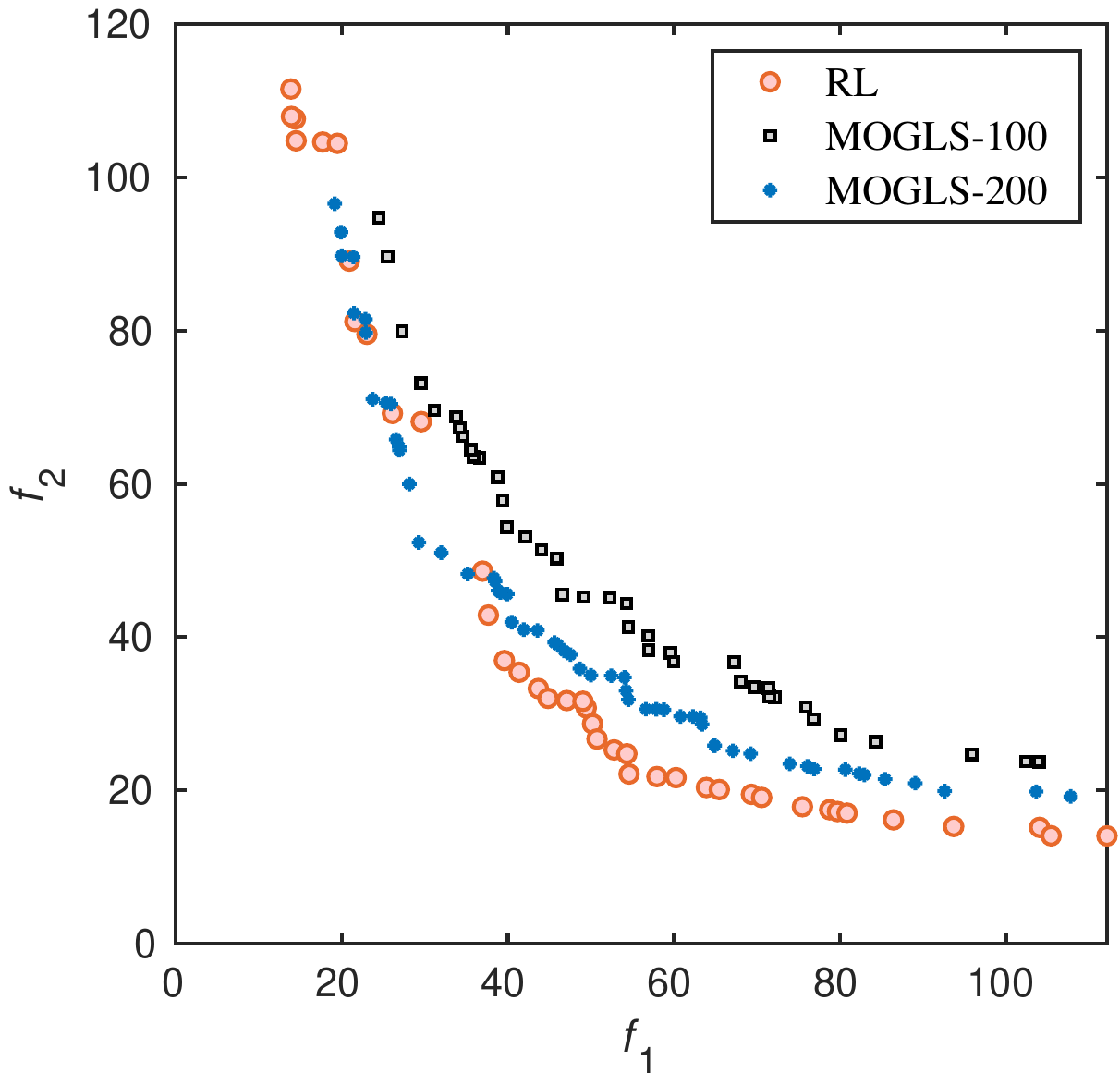}%
	}
	\caption{PFs obtained by MOGLS-100, MOGLS-200 and DRL-MOA on 200-city bi-objective TSP instances.}
	\label{fig:ls}
\end{figure}

Since the local search can further improve the quality of solutions obtained by DRL as reported in \cite{deudon2018learning}, we thus use a simple 2-opt local search to post-process the solutions obtained by DRL-MOA, leading to the results of DRL-MOA+LS. It is noted that the 2-opt is conducted only once for each solution and it only costs several seconds in total. Thus results of the experiments on MOGLS-100, MOGLS-200, MOGLS-300, DRL-MOA and DRL-MOA+LS are presented in \tref{table:ls}. For clarity, only the PFs obtained by MOGLS-100, MOGLS-200 and our method are visualized in \fref{fig:ls}. 

It is found that using local search to post-process the solutions can further improve the performance. It can be observed that DRL-MOA+LS outperforms the compared MOGLS on all instances while requiring much less computation time. DRL-MOA without local search can also outperform MOGLS on all 200-city instances even the MOGLS has run for 1000 seconds, while DRL-MOA only requires about 13 seconds. Although MOGLS performs slightly better than DRL-MOA on 100-city instances, DRL-MOA can always obtain a comparable result within 7 seconds. 

Results of the experiments indicate the fast computing speed and guaranteed performance of the DRL-MOA method. Moreover, using local search to post-process the solutions can further improve the performance. 

\begin{table}[htbp]
  \centering
  \caption{HV values of the PFs obtained by MOGLS-100, MOGLS-200, MOGLS-300, DRL-MOA and DRL-MOA+LS. The best and the second vest HVs are marked in gray and light gray Background. The longest computation time is marked bold.}
    \begin{tabular}{clrrrr}
    \toprule
    \multicolumn{1}{l}{Instances} &       & \multicolumn{2}{c}{100-city} & \multicolumn{2}{c}{200-city} \\
    \midrule
    \multirow{6}[4]{*}{Mixed } &       & \multicolumn{1}{l}{HV} & \multicolumn{1}{l}{Time/s} & \multicolumn{1}{l}{HV} & \multicolumn{1}{l}{Time/s} \\
\cmidrule{2-6}          & MOGLS-100 & 1848  & 143.3 & 5800  & 394.6 \\
          & MOGLS-200 & 1986  & 254.9 & 6516  & 738.8 \\
          & MOGLS-300 & \cellcolor[rgb]{ .851,  .851,  .851}2037  & \textbf{349.8} & 6794  & \textbf{1073.7} \\
          & DRL-MOA & 2022  & 6.6   & \cellcolor[rgb]{ .851,  .851,  .851}6911  & 12.9 \\
          & DRL-MOA+LS & \cellcolor[rgb]{ .651,  .651,  .651}2073  & 12.7  & \cellcolor[rgb]{ .651,  .651,  .651}6988  & 23.2 \\
    \midrule
    \multirow{6}[4]{*}{Euclidean } &       & \multicolumn{1}{l}{HV} & \multicolumn{1}{l}{Time/s} & \multicolumn{1}{l}{HV} & \multicolumn{1}{l}{Time/s} \\
\cmidrule{2-6}          & MOGLS-100 & 3127  & 138.1 & 13799  & 374.8 \\
          & MOGLS-200 & 3362  & 246.3 & 15093  & 697.1 \\
          & MOGLS-300 & \cellcolor[rgb]{ .851,  .851,  .851}3450  & \textbf{348.1} & 15716  & \textbf{1005.6} \\
          & DRL-MOA & 3342  & 6.4   & \cellcolor[rgb]{ .851,  .851,  .851}15750  & 12.9 \\
          & DRL-MOA+LS & \cellcolor[rgb]{ .651,  .651,  .651}3474  & 12.4  & \cellcolor[rgb]{ .651,  .651,  .651}16117  & 24.1 \\
    \bottomrule
    \end{tabular}%
  \label{table:ls}%
\end{table}%

\subsection{Effectiveness of parameter-transfer strategy}
\begin{figure}[htbp]
	\centering
	\subfloat[Performances of two models: one is trained via the parameter-transfer strategy; another is trained without transferring the network weights. They are both trained  on 120,000 instances for 5 epochs.]{\includegraphics[width=1.5in]{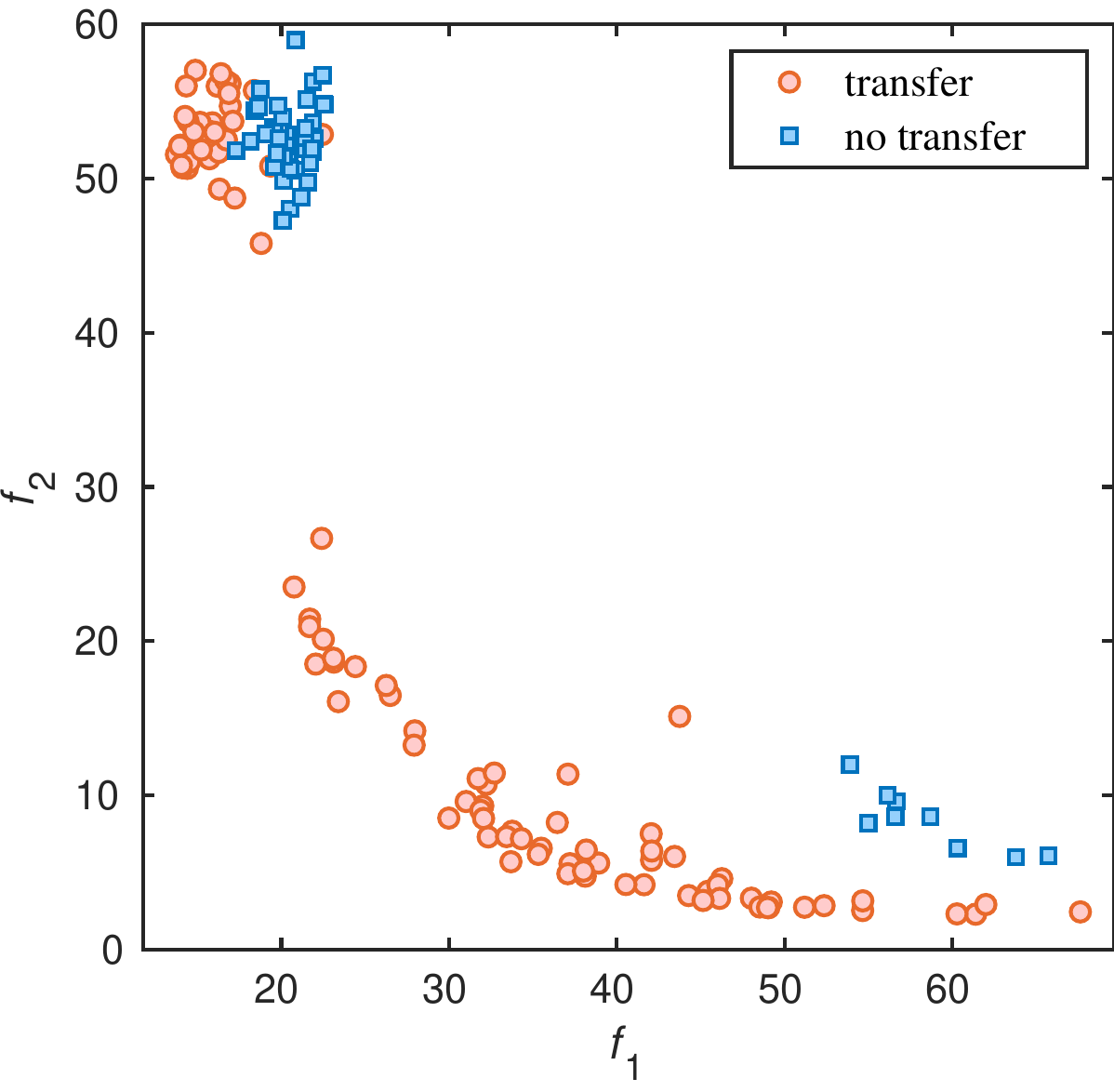}%
	}
	\hfil
	\subfloat[Performances of two models: the first model is the same with that in (a); the second model is trained on 240,000 instances for 10 epochs without applying the parameter-transfer strategy.   ]{\includegraphics[width=1.5in]{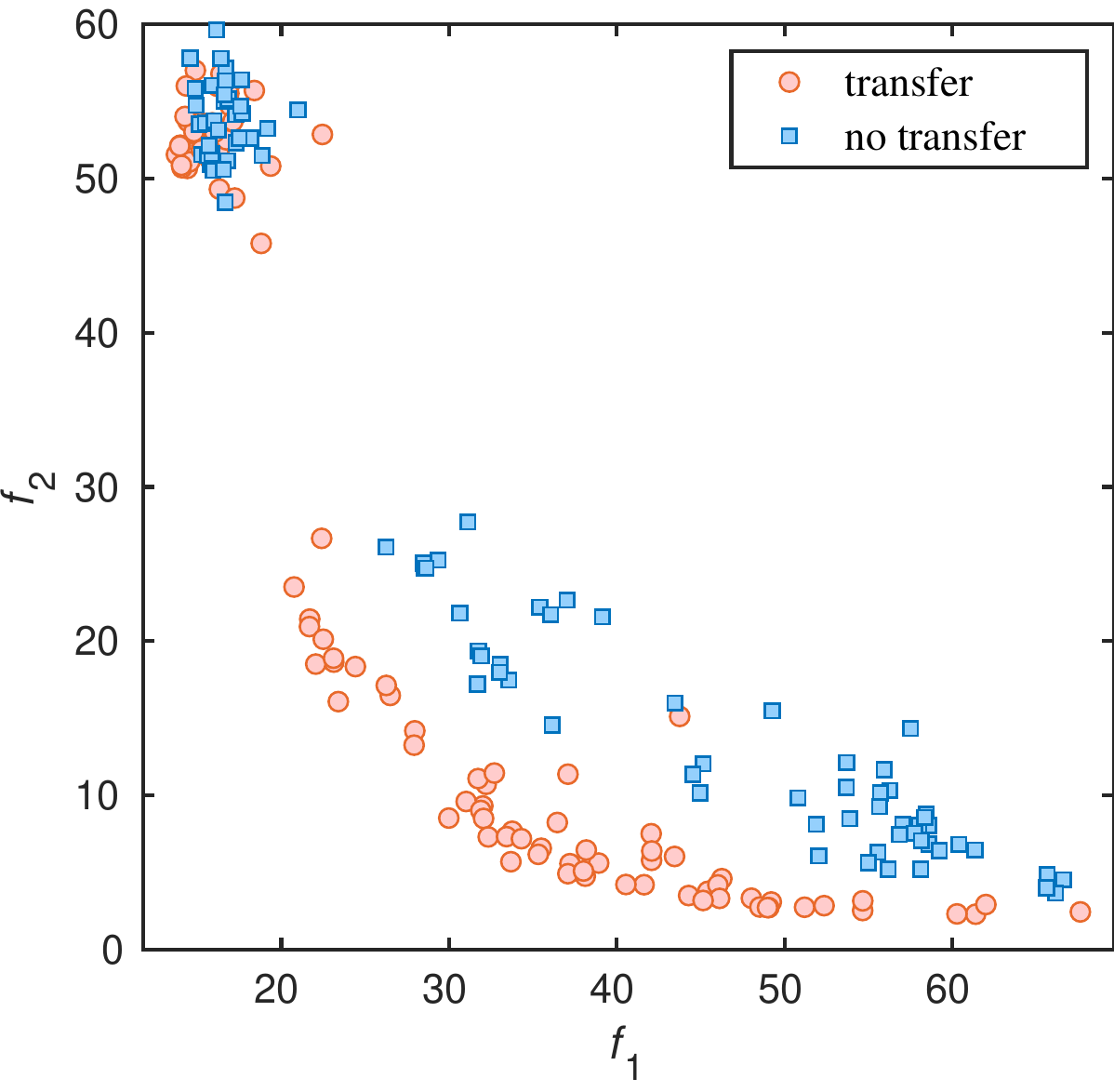}%
	}
	\caption{Performances of the models that are trained with and without the parameter-transfer strategy.}
	\label{transfer}
\end{figure}

In this section, the effectiveness of the neighborhood-based parameter-transfer strategy is checked experimentally. At first, performances of the models that are trained with and without the parameter-transfer strategy are compared. They are both trained on 120,000 20-city MOTSP instances for five epochs. PFs obtained by the two models are presented in \fref{transfer} (a). It is obvious that performance of the model is dramatically poor if the parameter-transfer strategy is not used for its training.

Moreover, we train the model without applying the parameter-transfer strategy on 240,000 instances for 10 epochs, that is, the model is trained four times longer than before. The result is presented in \fref{transfer} (b). It can be seen that, without the parameter-transfer strategy, even if the model is trained four times longer, it still exhibits a poor performance. Thus it is effective and efficient to apply the parameter-transfer strategy for training; otherwise it is impossible to obtain a promising model within a reasonable time. 

\subsection{The impact of training on different number of cities}
\begin{figure}[htbp]
\centering
\subfloat[Model trained on 20-city instances]{\includegraphics[width=1.5in]{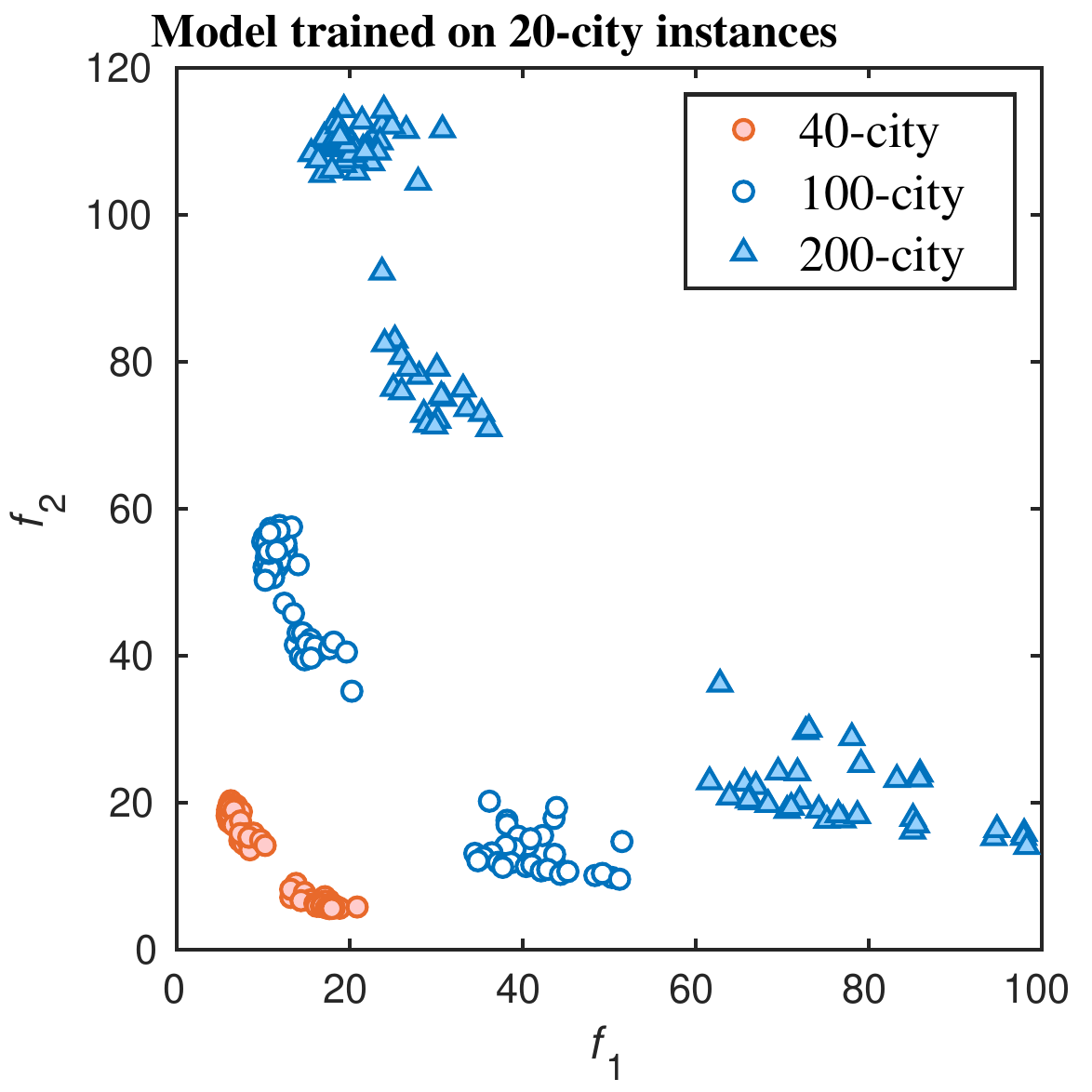}%
}
\hfil
\subfloat[Model trained on 40-city instances]{\includegraphics[width=1.5in]{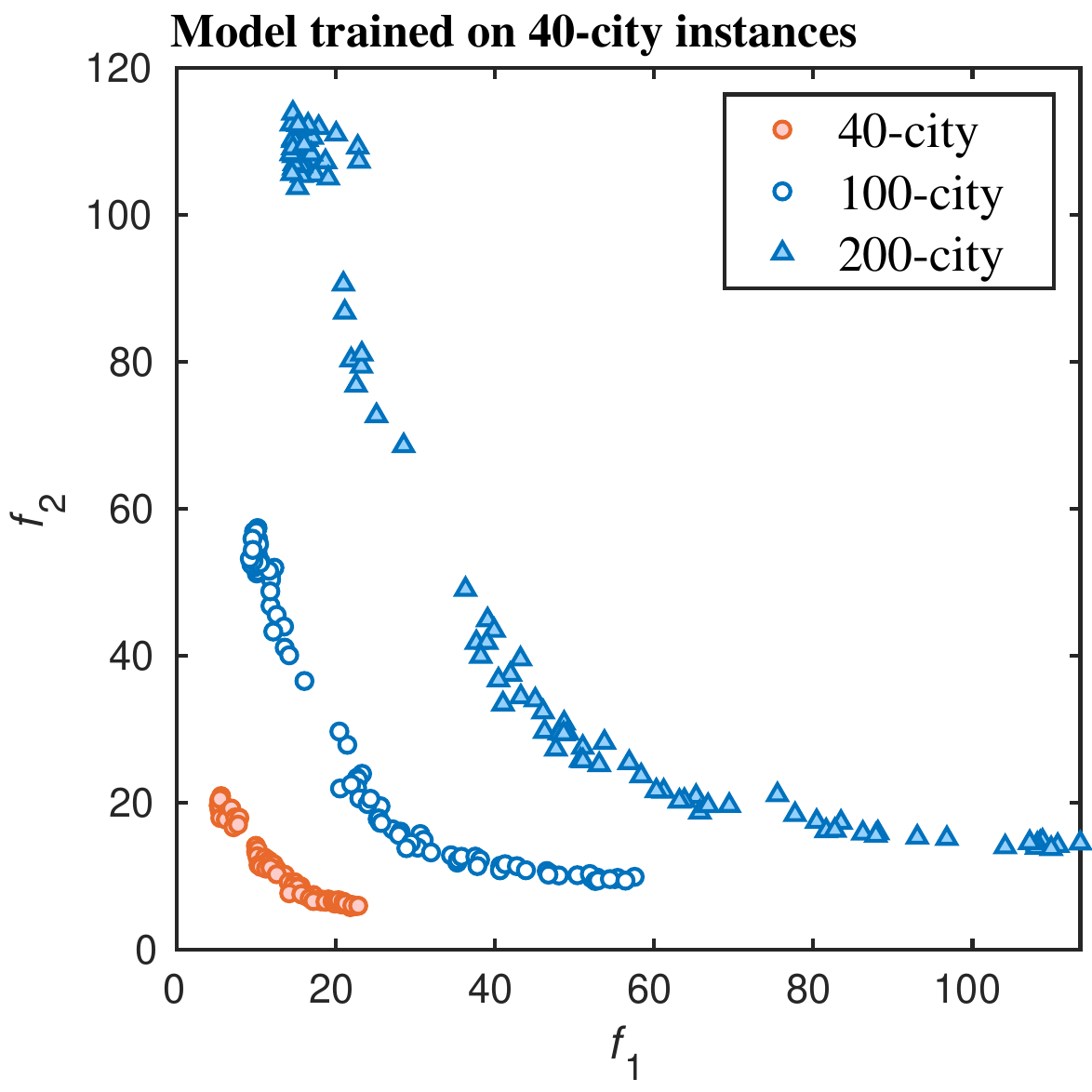}%
}
\caption{The two models trained respectively on 20- and 40-city \textbf{Euclidean bi-objective TSP instances}. They are used to approximate the PF of 40-, 100-, 200-city problems.}
\label{4_static_compare}
\end{figure}

The forgoing models are trained on 40-city instances. In this part, we try to figure out whether there is any difference of the model performance if the model is trained on 20-city instances. Performance of the model that is trained on 20-city instances is presented in \fref{4_static_compare} (a).

It can be observed that the model trained on 20-city instances exhibits an apparently worse performance than the one trained on 40-city instances. A large number of solutions obtained by the 20-city model are crowded in several regions and there are less non-dominated solutions. A possible reason for the deteriorated result is that, when training on 40-city instances, 40 city selecting decisions are made and evaluated in the process of training each instance, which are twice of that when training on 20-city instances.
Loosely speaking, if both the two models use $120,000$ instances, the 40-city model are trained based on $120,000 \times 40$ cities which are twice of that of 20-city model. Therefore, the model trained on 40-city instances is better. And we can simply increase the number of training instances to improve the performance. 

Lastly, it is also interesting to see that the solutions output by DRL-MOA are not all non-dominated. Moreover, these solutions are not distributed evenly (being along with the provided search directions). These issues deserve more studies in future.


\subsection{Summary of the results}
Observed from the experimental results, we can conclude that the DRL-MOA is able to handle MOTSP both effectively and efficiently. In comparison with classical multi-objective optimization methods, DRL-MOA has shown some encouragingly new characteristics, e.g., strong generalization ability, fast solving speed and promising quality of the solutions, which can be summarized as follows.

\begin{itemize}
	\item Strong generalization ability. Once the trained model is available, it can scale to newly encountered problems with no need of retraining the model. Moreover, its performance is less affected by the increase of number of cities compared to existing methods.
	\item A better balance between the solving speed and the quality of solutions. The Pareto optimal solutions can be always obtained within a reasonable time while the quality of the solutions is still guaranteed. 
\end{itemize}

\section{Conclusion}
Multi-objective optimization, appeared in various disciplines, is a fundamental mathematical problem. Evolutionary algorithms have been recognized as suitable methods to handle such problem for a long time. However, evolutionary algorithms, as iteration-based solvers, are difficult to be used for on-line optimization. Moreover, without the use of a large number of iterations and/or a large population size, evolutionary algorithms can hardly solve large-scale optimization problems \cite{zhang2016adecision,ming2019emo,lust2010multiobjective}.

Inspired by the very recent work of Deep Reinforcement Learning (DRL) for single-objective optimization, this study provides a new way of solving the multi-objective optimization problems by means of DRL and has found very encouraging results. In specific, on MOTSP instances, the proposed DRL-MOA significantly outperforms NSGA-II, MOEA/D and MOGLS in terms of the solution convergence, spread performance as well as the computing time, and thus, making a strong claim to use the DRL-MOA, a non-iterative solver, to deal with MOPs in future. 

With respect to the future studies, first in the current DRL-MOA, a 1-D convolution layer which corresponds to the city information is used as inputs. Effectively, a distance matrix used as inputs can be further studied, i.e., using a 2-D convolution layer. Second, the distribution of the solutions obtained by the DRL-MOA are not as even as expected. Therefore, it is worth investigating how to improve the distribution of the obtained solutions. Overall, multi-objective optimization by DRL is still in its infancy. It is expected that this study will motivate more researchers to investigate this promising direction, developing more advanced methods in future.


%





\ifCLASSOPTIONcaptionsoff
  \newpage
\fi



\bibliographystyle{IEEEtran}
\bibliography{mybib}
%



%

\begin{IEEEbiography}[{\includegraphics[width=1in,height=1.25in,clip,keepaspectratio]{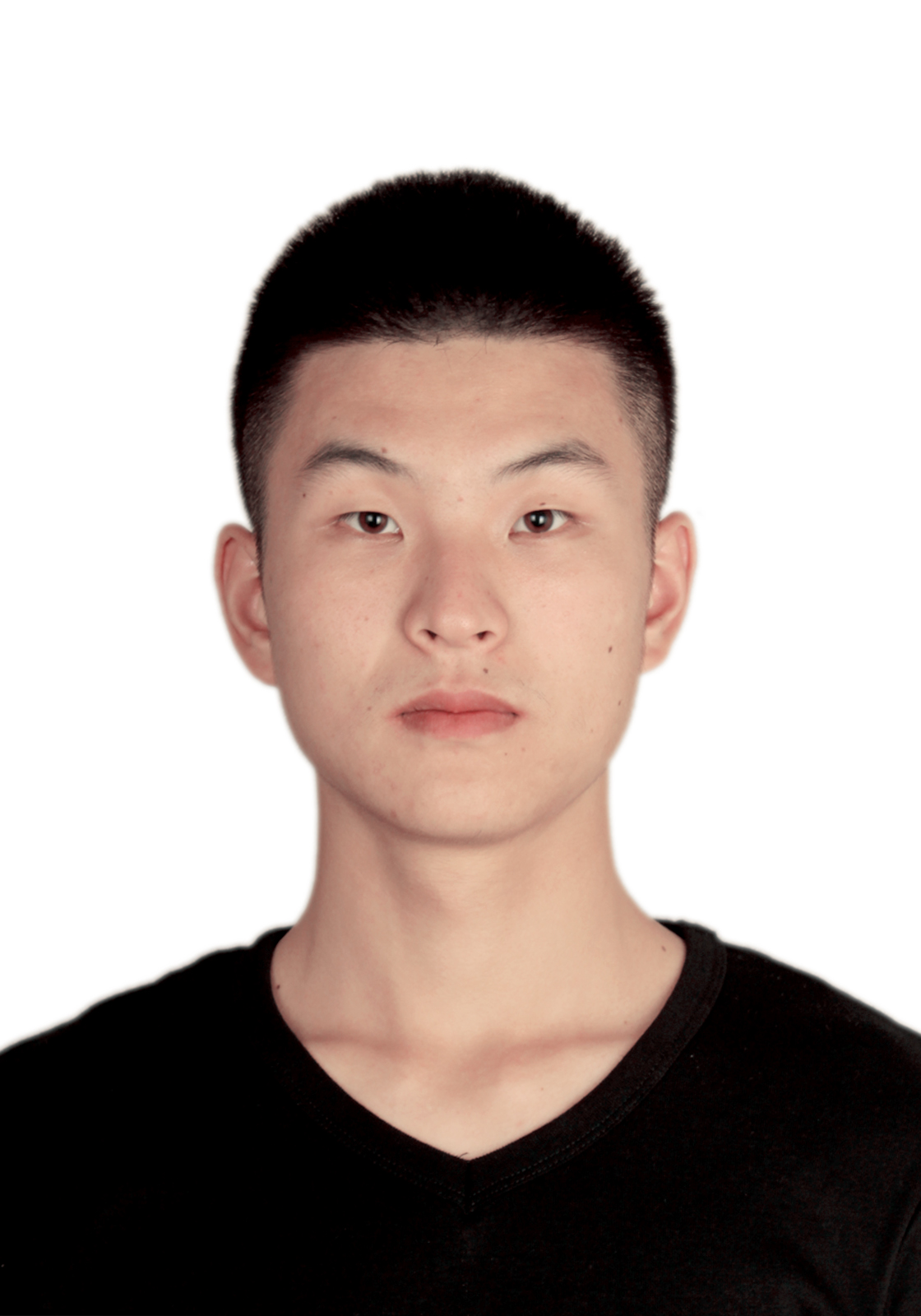}}]{Kaiwen Li} received the B.S., M.S. degrees from National University of Defense Technology (NUDT), Changsha, China, in 2016 and 2018.
	
He is a student with the College of Systems Engineering, NUDT. His research interests include prediction technique, multiobjective optimization, reinforcement learning, data mining, and optimization methods on Energy Internet.

\end{IEEEbiography}

\begin{IEEEbiography}[{\includegraphics[width=1in,height=1.25in,clip,keepaspectratio]{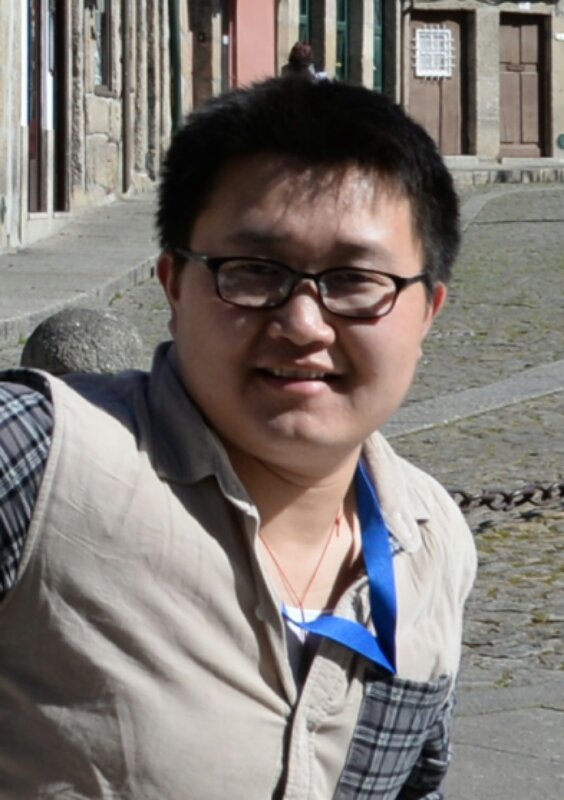}}]{Rui Wang} received the B.S. degree from National University of Defense Technology (NUDT), Changsha, China, in 2008 and the Ph.D. degree from University of Sheffield, Sheffield, U.K., in 2013.
	
He is a Lecturer with the College of Systems Engineering, NUDT. His research interests include evolutionary computation, multiobjective optimization, machine learning, and various applications using evolutionary algorithms.
\end{IEEEbiography}

\begin{IEEEbiography}[{\includegraphics[width=1in,height=1.25in,clip,keepaspectratio]{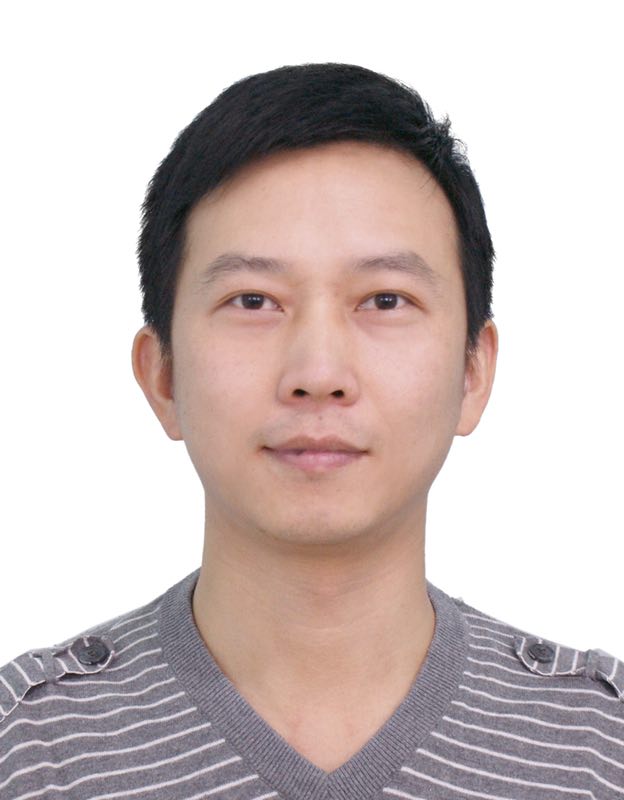}}]{Tao Zhang} received the B.S., M.S., Ph.D. degrees from National University of Defense Technology (NUDT), Changsha, China, in 1998, 2001, and 2004, respectively.
	
He is a Professor with the College of Systems Engineering, NUDT. His research interests include multicriteria decision making, optimal scheduling, data mining, and optimization methods on energy Internet network.
\end{IEEEbiography}




\end{document}